\documentclass[lettersize,journal]{IEEEtran}
\usepackage{amssymb}  
\usepackage{amsmath,amsfonts}
\usepackage{algorithmic}
\usepackage{array}
\usepackage[caption=false,font=normalsize,labelfont=sf,textfont=sf]{subfig}
\usepackage{textcomp}
\usepackage{stfloats}
\usepackage[font=footnotesize]{caption}
\usepackage{url}
\usepackage{booktabs} 
\usepackage{graphicx}
\usepackage{float}

\usepackage[hidelinks]{hyperref}
\def\BibTeX{{\rm B\kern-.05em{\sc i\kern-.025em b}\kern-.08em
    T\kern-.1667em\lower.7ex\hbox{E}\kern-.125emX}}
\usepackage{balance}

\usepackage[markup=final]{changes}
\definecolor{deepgreen}{rgb}{0.0, 0.5, 0.0}
\definecolor{white}{rgb}{0.0, 0.0, 0.0}
\definechangesauthor[color=red]{R1}
\definechangesauthor[color=deepgreen]{R2}

\usepackage{todonotes}
\setcommentmarkup{\todo[color={authorcolor!20},size=\scriptsize]{#3: #1}}
\makeatletter
\renewcommand{\added}[2][]{#2}   
\makeatother

\begin{document}
\title{IRSAMap:Towards Large-Scale, High-Resolution Land Cover Map Vectorization}
\author{Yu Meng, Ligao Deng, Zhihao Xi, Jiansheng Chen, Jingbo Chen, Anzhi Yue, Diyou Liu, Kai Li, \IEEEmembership{Student Member,~IEEE}, Chenhao Wang, Kaiyu Li, Yupeng Deng*, and Xian Sun, \IEEEmembership{Senior Member,~IEEE}
\thanks{This work has been accepted for publication in IEEE Transactions on Geoscience and Remote Sensing.
The final version is available at \href{https://doi.org/10.1109/TGRS.2025.3600249}{DOI: 10.1109/TGRS.2025.3600249}.
  
This research was funded by the National Key R\&D Program of China under grant number 2021YFB3900503. (Corresponding author: Yupeng Deng.)

Yu Meng, Zhihao Xi, Jiansheng Chen, Jingbo Chen, Anzhi Yue, Diyou Liu, Yupeng Deng and Xian Sun are with Aerospace Information Research Institute, Chinese Academy of Sciences, Beijing 100094, China.(e-mail: mengyu@aircas.ac.cn; xizh@aircas.ac.cn; chenjs@aircas.ac.cn; chenjb@aircas.ac.cn; yueaz@aircas.ac.cn; liudiyou@aircas.ac.cn; dengyp@aircas.ac.cn; sunxian@aircas.ac.cn)

Ligao Deng, Kai Li and Chenhao Wang are with Aerospace Information Research Institute,
Chinese Academy of Sciences, Beijing 100094, China, and also with School of Electronic, Electrical and Communication Engineering, University of Chinese Academy of Sciences, Beijing 100049, China.(e-mail: dengligao22@mails.ucas.ac.cn; likai211@mails.ucas.ac.cn; wangchenhao22@mails.ucas.ac.cn)

Kaiyu Li is with the School of Software Engineering, Xi'an Jiaotong University, Xi'an 710049, China. (e-mails: likyoo.ai@gmail.com)
}}

\maketitle

\begin{abstract}
With the continuous enhancement of remote sensing image resolution and the rapid advancement of deep learning techniques, land cover mapping is undergoing a significant transformation from pixel-level segmentation to object-based vector modeling. This shift imposes higher demands on deep learning models, requiring not only precise delineation of object boundaries but also the preservation of topological consistency among geographic elements. However, existing public datasets face three major limitations: limited class annotations, restricted data scale, and the lack of spatial structural information, which severely hinder the development of breakthrough methods in high-resolution remote sensing vectorization. To address these challenges, we present IRSAMap, the first global remote sensing dataset designed for large-scale, high-resolution, multi-feature land cover vector mapping. This dataset offers four key advantages: First, a comprehensive element vector annotation system that includes over 1.8 million instances of 10 typical natural and man-made objects, such as buildings, roads, rivers, and trees, employing a unified vector annotation standard framework that ensures both semantic integrity and spatial structural accuracy. Second, an intelligent annotation workflow incorporating “manual pre-annotation + AI-based training and inference + manual review and correction,” which enhances annotation efficiency while ensuring consistency. Third, a global coverage that spans 79 regions across six continents, representing diverse terrain types, including urban and rural areas, with a total coverage area exceeding 1,000 square kilometers. Fourth, multi-task adaptability, supporting various tasks such as pixel-level land cover classification, building outline regularization extraction, road centerline extraction, and panoramic segmentation. As a fundamental resource for remote sensing intelligent interpretation, IRSAMap provides a standardized benchmark for the paradigm shift from pixels to objects, which will significantly advance the development of high-precision geographic feature automation, collaborative modeling, and other cutting-edge research directions. The dataset is of great value for applications such as global geographic information updating and digital twin construction. IRSAMap is publicly available at https://github.com/ucas-dlg/IRSAMap.
\end{abstract}


\section{Introduction}
\IEEEPARstart{T}{he} rapid advancement of high-resolution remote sensing technologies has provided unprecedented data support for the fine-scale identification of land cover. Automatic land cover classification, as a core task in remote sensing intelligent interpretation, directly contributes to critical national needs such as urban planning~\cite{jhawar2013urban}, environmental monitoring~\cite{das2022land}, and disaster assessment~\cite{zhang2024event}. Traditional manual interpretation methods are increasingly inadequate for handling the vast volume of remote sensing data. Therefore, there is an urgent need to develop automated and intelligent methods for land cover extraction.

\begin{figure*}[htbp]
  \centering
  \includegraphics[width=\textwidth]{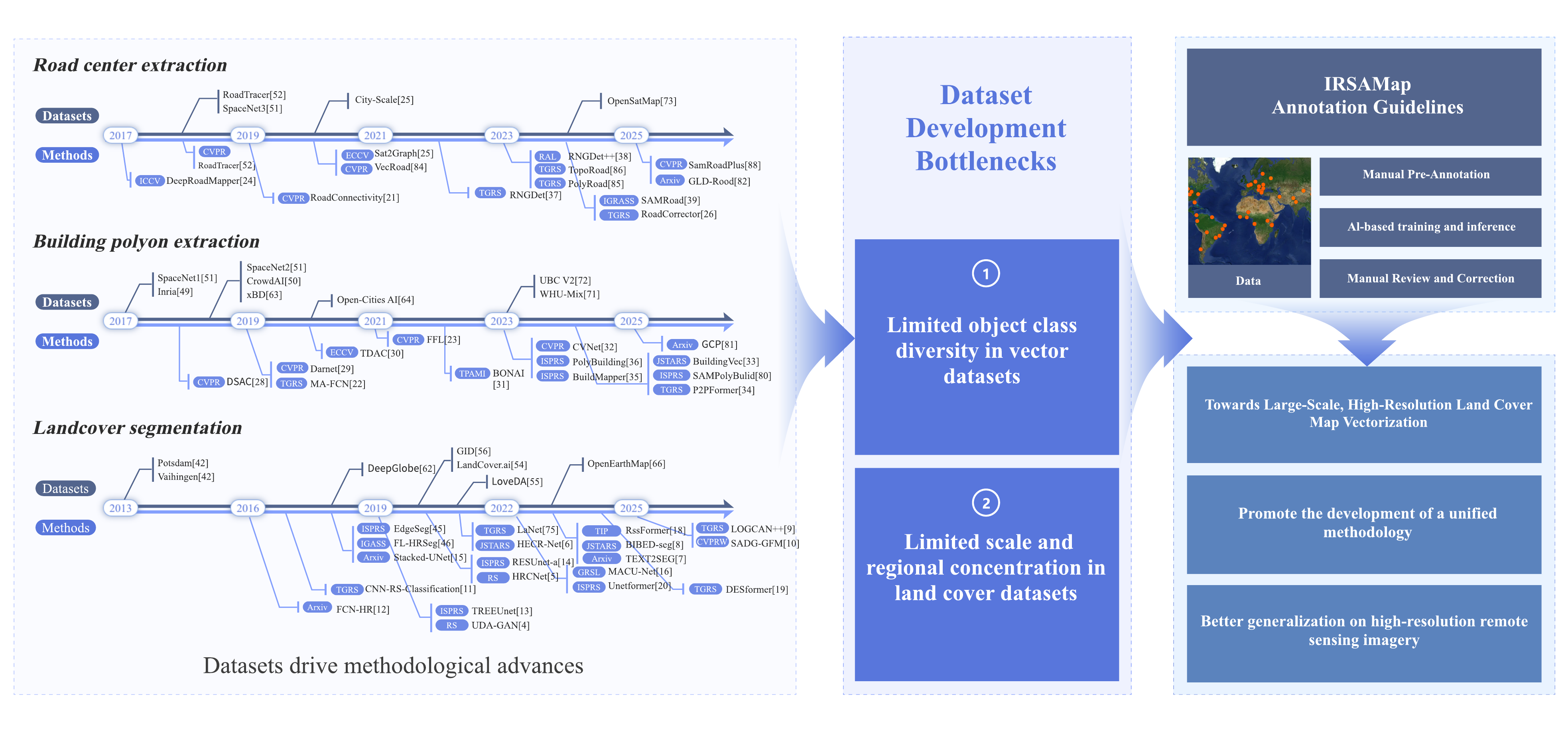} 
  \captionsetup{justification=centering}  
  \caption{Development trajectory of geographic element extraction methods and an overview of dataset bottlenecks.}
  \label{fig:intro}
\end{figure*}

The emergence of deep learning methods has revolutionized pixel-level automatic classification for land cover. Since 2016, researchers have progressively explored land cover extraction models based on semantic segmentation, aiming at automated land cover mapping\added[id=R1,comment={1}]{~\cite{benjdira2019unsupervised, xu2020hrcnet,liu2021hecr, zhang2023text2seg, 10018433, ma2025logcan++, yaghmour2025sensor}}. Emmanuel Maggiori et al.~\cite{maggiori2017convolutional} proposed an end-to-end pixel classification framework for remote sensing images based on fully convolutional networks (FCNs), utilizing two-stage training and multi-scale modules, achieving superior object segmentation accuracy compared to traditional SVM classifiers. Similarly, Jamie Sherrah et al.~\cite{sherrah2016fully} employed FCNs to obtain more precise object boundaries on the Vaihingen and Potsdam datasets. Concurrently, U-Net based land cover models have also demonstrated promising performance.\added[id=R1,comment={1}]{~\cite{yue2019treeunet, diakogiannis2020resunet}}. Andrew Khalel et al.~\cite{khalel2018automatic} achieved better results on the Inria and Massachusetts datasets by concatenating U-Net models, while the MACU-Net~\cite{9343296} network, which incorporates multi-scale skip connections and asymmetric convolutions on the U-Net architecture, demonstrated outstanding performance on large-scale land cover classification datasets. With the impressive performance of Transformer-based models in computer vision, land cover pixel-level classification methods based on Transformers~\cite{vaswani2017attention} have garnered attention\added[id=R1,comment={1}]{~\cite{xu2023rssformer, liu2024desformer}}. The UnetFormer model~\cite{wang2022unetformer}, which integrates a CNN-based encoder and a Transformer-based decoder, combines the advantages of both architectures. Methods in this phase predominantly focused on improving boundary clarity and class discriminability in segmentation results, with outputs represented as rasterized mask images. However, as remote sensing applications increasingly demand high spatial structure, topological relationships, and semantic integrity, the limitations of rasterized representations in terms of accuracy and spatial relationships have become more apparent.

With the advancement of land cover extraction research and the expansion of algorithmic applications in operational scenarios, researchers have identified limitations in semantic segmentation algorithms, particularly when handling complex boundaries. This is especially true for regular man-made features, such as buildings and roads, where the topological structure remains incomplete and is difficult to edit. To address the shortcomings of traditional raster-based mask representations in spatial geometry modeling, researchers have begun exploring the conversion of segmentation results into vector formats\added[id=R1,comment={1}]{~\cite{batra2019improved, wei2019toward}}. This approach, combined with post-processing techniques, enables the extraction of polygonal boundaries or skeletal structures, thus facilitating object-level representation of land cover. Man-made features like buildings and roads, which exhibit strong geometric characteristics and structural regularity, have become primary targets for vectorization extraction. Early approaches to building and road vectorization largely relied on post-processing segmentation results to generate vectorized outputs. For example, Nicolas Girard et al.~\cite{girard2021polygonal} introduced a frame field prediction branch in semantic segmentation networks to represent local principal directions, aligning the frame field directions with building segmentation edges to simplify and form vectorized building outlines. DeepRoadMapper~\cite{mattyus2017deeproadmapper} extracted road segmentation lines, then refined the road centerline and used shortest path algorithms to connect fragmented roads, forming the final road network. Sat2Graph~\cite{he2020sat2graph} employed tensor encoding to represent road network node positions and directions, followed by post-processing methods for node connectivity. However, these methods treated semantic segmentation and post-processing as two independent stages, leading to suboptimal vectorization accuracy due to the inability to jointly optimize both phases. In recent years, end-to-end approaches for vectorizing man-made feature extraction have been proposed\added[id=R1,comment={1}]{~\cite{li2024roadcorrector, yin2025towards, marcos2018learning, cheng2019darnet, hatamizadeh2020end, wang2022learning, xu2022cvnet, huang2024instance, zhang2024p2pformer}}. BuildMapper~\cite{wei2023buildmapper} introduced a learnable, end-to-end building contour framework, replacing complex post-processing with a contour evolution module, directly outputting regularized vectorized building outlines. PolyBuilding~\cite{hu2023polybuilding} utilized a Transformer architecture to simultaneously output building vertices and bounding boxes, achieving end-to-end training of building vectorization, significantly improving contour regularity and accuracy. RNGDet~\cite{xu2022rngdet} and RNGDet++~\cite{xu2023rngdet++} adopted imitation learning techniques, iteratively retrieving road nodes during training to enable end-to-end road network training. SamRoad~\cite{hetang2024segment} designed a road node connection module after the SAM~\cite{kirillov2023segment} encoder extracted image features, replacing complex road network node connection post-processing algorithms. The aforementioned vectorization methods focus on extracting a single land class. TopDiG~\cite{yang2023topdig} proposed a universal remote sensing image topology extraction framework that can extract both polygonal and linear topological structures. However, due to the lack of vector label datasets, this method has only been tested on road network and building vector extraction results on different datasets. The findings of these studies indicate that land cover research is progressively transitioning from raster-based outputs and pixel-level segmentation to object-level vectorized results.

The advancement of land cover algorithms is fundamentally supported by data, with each algorithmic innovation relying on data as its foundation. In the early stages of deep learning, researchers transferred methods from natural domains to remote sensing, resulting in the use of mask-based annotation formats similar to those in natural domains. The introduction of the ISPRS Potsdam/Vaihingen dataset in 2014~\cite{rottensteiner2014isprs}, which features multi-class segmentation, significantly advanced the development of remote sensing image segmentation algorithm~\cite{volpi2016dense, kampffmeyer2016semantic, marmanis2018classification, doi2018effect, zhang2018vprs, chen2018symmetrical}. However, as research progressed, it became evident that existing methods, which typically generate raster predictions via models followed by post-processing to convert them into vector data, often resulted in final vector outputs with topological accuracy insufficient to meet practical demands, particularly for regular man-made features like roads and buildings. Consequently, vector-labeled datasets such as Inria~\cite{maggiori2017can}, CrowdAI~\cite{mohanty2018crowdai}, Spacenet~\cite{van2018spacenet}, and City-Scale~\cite{he2020sat2graph} were introduced, further driving the development of vector extraction methods for man-made features~\cite{mattyus2017deeproadmapper, he2020sat2graph, wei2023buildmapper, hu2023polybuilding, marcos2018learning, bastani2018roadtracer, xu2022rngdet, jiao2024polyr}. Simultaneously, large-scale high-resolution multi-class segmentation datasets, including LandCover.ai~\cite{boguszewski2021landcover}, LoveDA~\cite{wang2loveda}, GID~\cite{tong2020land}, and FBP~\cite{tong2023enabling}, have propelled the transition of multi-class segmentation algorithms to large-scale applications~\cite{9343296, wang2021transformer, wang2022unetformer, chen2024coarse, guo2024building, lu2024multi}. However, these datasets often feature raster-based annotations with insufficient precision, limiting the development of land cover models for large-scale applications. As indicated by the above analysis and~\autoref{fig:intro}, current datasets face two main development bottlenecks: First, vector datasets are typically limited to a single land class, such as building outlines or road centerlines, and lack a unified multi-class semantic system or the ability to represent structures across classes. Second, most mainstream multi-class datasets still rely on raster annotations, which are suitable for semantic segmentation but insufficient for vector modeling due to limitations in geometric accuracy and topological consistency. This mismatch between available datasets and research methods has significantly hindered the innovative development of vectorization algorithms.

To advance the research of large-scale land cover vectorization methods, this paper introduces the IRSAMap, the first large-scale high-resolution land cover vector dataset. The dataset encompasses a variety of typical natural and man-made land cover classes, with all elements represented in a unified vector format. This annotation approach not only enhances the precision of spatial boundaries and structural representation while maintaining semantic diversity, but also provides a standardized data foundation for the automated vector extraction of multi-class collaborative representations. IRSAMap offers the following four key advantages:

1.Comprehensive Element Vector Annotation System: The IRSAMap dataset includes ten typical land cover classes, such as buildings, roads, water bodies, and trees. The annotation process adheres strictly to the "what you see is what you get" principle, with detailed delineations based on specific objects to ensure consistency in labeling throughout the large-scale data annotation process.

2.Intelligent Annotation Workflow: IRSAMap introduces a three-stage iterative annotation process: "manual pre-annotation + AI-based training and inference + manual review and correction." This workflow combines the strengths of both human annotation and AI, improving annotation efficiency while addressing common ambiguities such as building individual units and road widths. The iterative process continuously refines consistency with each annotation cycle.

3.Extensive and Diverse Annotation: The dataset covers 79 representative regions across six continents, with a total annotation area of 1,024 square kilometers, including 1.8 million land cover instances. The dataset spans both urban and rural areas, encompassing a wide range of geographical and landform types. This large-scale image annotation and diverse geographical distribution make IRSAMap highly suitable for large-scale vectorization and mapping applications. 

4.Multi-task Adaptation Design: IRSAMap is compatible with three task paradigms: pixel-level tasks (such as semantic segmentation and panoptic segmentation), object-level tasks (such as instance segmentation and building footprint extraction), and vector extraction tasks (such as regularization of polygonal features and centerline extraction of linear features).

\section{Related Work}
\subsection{Raster-annotated Dataset}
\begin{table}[ht]
  \centering
  \caption{Overview of the Raster-annotated Dataset, where "b" refers to building extraction, "RA" to road surface extraction, "RC" to road centerline, and "LC" to land cover.}
  \resizebox{\columnwidth}{!}{
  \begin{tabular}{clcccccc}
  \hline
  \textbf{Resolution} & \textbf{Dataset} & \textbf{Year} & \textbf{Label Type} & \multicolumn{4}{c}{\textbf{Task}} \\ \hline
                       &                  &                &                    & \textbf{B} & \textbf{RA} & \textbf{RC} & \textbf{LC} \\ \hline
  0.05  & ISPRS Potsdam~\cite{rottensteiner2014isprs} & 2013 & Raster & \checkmark & \checkmark & $\times$ & \checkmark \\
  0.09  & ISPRS Vaihingen~\cite{rottensteiner2014isprs} & 2013  & Raster  & \checkmark & \checkmark & $\times$  & \checkmark  \\ 
  0.3   & Inria~\cite{maggiori2017can}           & 2017  & Raster  & \checkmark & $\times$     & $\times$  & $\times$  \\ 
  0.3-0.5 & DeepGlobe~\cite{8575485}      & 2018  & Raster  & \checkmark & \checkmark     & $\times$  & \checkmark  \\ 
  0.3-0.8 & xBD~\cite{weber2020building}            & 2019  & Raster  & \checkmark     & $\times$     & $\times$  & $\times$  \\ 
  0.02-0.2 & Open-Cities AI~\cite{gfdrr2020} & 2020  & Raster  & \checkmark & $\times$     & $\times$  & $\times$ \\ 
  4     & GID~\cite{tong2020land}             & 2020  & Raster  & $\times$     & $\times$     & $\times$ & \checkmark \\ 
  0.25  & LandCover.ai~\cite{boguszewski2021landcover}    & 2020  & Raster  & \checkmark & \checkmark & $\times$  & \checkmark \\ 
  0.3   & LoveDA~\cite{wang2loveda}          & 2021  & Raster  & \checkmark & \checkmark & $\times$  & \checkmark  \\ 
  0.5   & CHN6-CUG~\cite{zhu2021global}         & 2021  & Raster  & $\times$     & $\times$ & \checkmark  & $\times$ \\ 
  0.25-0.5 & OpenEarthMap~\cite{xia2023openearthmap}   & 2023  & Raster  & \checkmark & \checkmark & $\times$  & \checkmark  \\
  \hline
  \end{tabular}
  }
  \label{tab:tab1}
\end{table}

A number of datasets have been developed for multi-class land-cover extraction tasks using raster-based annotations. Representative raster datasets commonly used in recent studies are summarized in~\autoref{tab:tab1}. As shown in the~\autoref{tab:tab1}, although most datasets provide high-resolution imagery, the majority do not include annotations for road centerlines. In cases where road features are labeled, the annotations typically represent road surfaces rather than precise centerlines. Early research predominantly employed datasets with medium to low resolution imagery, such as MCD12Q1~\cite{friedl2010modis}, the National Land Cover Database (NLCD)~\cite{homer2004development}, GlobeLand30~\cite{chen2015global}, and LandCoverNet~\cite{alemohammad2020landcovernet}. These datasets have significantly contributed to the application of deep learning in semantic segmentation of remote sensing imagery. However, due to the limited resolution, they fall short of meeting the current demands for fine-grained land cover extraction. With advances in imaging and data transmission technologies, higher-resolution datasets have been released, focusing on different task-specific scenarios. The International Society for Photogrammetry and Remote Sensing (ISPRS) released the Potsdam and Vaihingen high-resolution datasets~\cite{rottensteiner2014isprs}, with imagery focused on urban and suburban areas. In contrast, the DeepGlobe~\cite{8575485} and LandCover.ai~\cite{boguszewski2021landcover} datasets are centered on rural regions, with a primary focus on rural remote sensing imagery. The GID dataset~\cite{tong2020land} consists of 1-meter resolution imagery from the Gaofen-2 satellite, covering multiple cities across China. The LoveDA dataset~\cite{wang2loveda} includes both urban and rural scenes to address accuracy degradation caused by scene differences. The OpenEarthMap dataset~\cite{xia2023openearthmap} provides annotations on remote sensing imagery from 97 regions across 44 countries, further enhancing the regional diversity of segmentation datasets and offering high annotation precision, thereby providing a robust data foundation for multi-class land cover segmentation algorithms. Notably, all the aforementioned datasets employ raster-based labeling, where land cover is annotated at the pixel level with class information. While this approach facilitates the training of semantic segmentation models, it presents limitations in terms of representing land cover boundaries, maintaining geometric structures, and handling topological relationships.

\subsection{Vector-annotated Dataset}
\begin{table}[ht]
  \centering
  \caption{Overview of the Vector-annotated Dataset, where "b" refers to building extraction, "RA" to road surface extraction, "RC" to road centerline, and "LC" to land cover.}
  \resizebox{\columnwidth}{!}{
  \begin{tabular}{clcccccc}
  \hline
  \textbf{Resolution} & \textbf{Dataset} & \textbf{Year} & \textbf{Label Type} & \multicolumn{4}{c}{\textbf{Task}} \\ \hline
                       &                  &                &                    & \textbf{B} & \textbf{RA} & \textbf{RC} & \textbf{LC} \\ \hline
  0.5  & SpaceNet~\cite{van2018spacenet} & 2018 & Vector & \checkmark & $\times$ & \checkmark & $\times$ \\ 
  0.3  & CrowdAI~\cite{mohanty2018crowdai} & 2018  & Vector  & \checkmark & $\times$ & $\times$  & $\times$  \\ 
  0.6  & RoadTracer~\cite{bastani2018roadtracer} & 2018  & Vector  & $\times$ & $\times$ & \checkmark & $\times$ \\ 
  1    & City-Scale~\cite{he2020sat2graph} & 2020  & Vector  & $\times$ & $\times$ & \checkmark & $\times$ \\ 
  0.08-0.64   & WHU-Mix(vector)~\cite{luo2023diverse}  & 2023  & Vector  & \checkmark & $\times$ & $\times$  & $\times$ \\ 
  0.5-0.8 & UBC V2~\cite{huang2023urban} & 2023  & Vector  & \checkmark & $\times$ & $\times$  & $\times$  \\ 
  0.15-0.3   & OpenSatmap~\cite{zhao2024opensatmap} & 2024  & Vector  & $\times$ & $\times$ & \checkmark & $\times$ \\ 
  0.5 & IRSAMap & 2025  & Vector  & \checkmark & \checkmark     & \checkmark  & \checkmark  \\ 
  \hline
  \end{tabular}
  }
  \label{tab:tab2}
\end{table}
To address the limitations of raster-based annotation schemes, vector-based labeling has attracted increasing attention in recent years. Representative vector annotation datasets commonly used in current studies are summarized in~\autoref{tab:tab2}. As observed, existing vector datasets generally do not support land cover classification tasks. Instead, most of them are limited to the extraction of single object classes, primarily buildings or road networks. In the area of building footprint extraction, several vector annotation datasets have been released, including CrowdAI~\cite{mohanty2018crowdai}, SpaceNet~\cite{van2018spacenet}, and Inria~\cite{chen2018symmetrical}. Building on these resources, Muying Luo et al.~\cite{luo2023diverse} integrated existing building data, corrected labeling errors, and introduced new vector instances, leading to the creation of the WHU-Mix (vector) dataset, which contains approximately 754,000 building instances. This dataset aims to address issues related to insufficient diversity in building data and inconsistent labeling quality. In the field of road vectorization, several representative datasets have also been developed. For example, the RoadTracer dataset~\cite{bastani2018roadtracer} is based on OpenStreetMap (OSM)~\cite{haklay2008openstreetmap} road vector data and covers core urban areas in 40 cities globally, with a total labeling area of approximately 960 square kilometers. However, due to copyright restrictions on the original remote sensing imagery, the dataset acquisition process is relatively complex. The City-Scale dataset~\cite{he2020sat2graph}, which modifies OSM labels, was constructed to cover 720 square kilometers across 20 states in the United States, primarily focusing on urban core areas. A road vector dataset was also introduced in the SpaceNet Challenge~\cite{van2018spacenet}, which covers both urban and rural areas, increasing sample diversity. However, due to image size limitations, it still faces challenges in representing large-scale road networks with well-defined continuity. In summary, existing vector datasets still exhibit notable limitations in terms of coverage and land cover types. Most datasets focus solely on a single category (such as buildings or roads), and the vector labels for different categories often come from different image sources, lacking a unified spatial reference system. Furthermore, for other land cover types, such as water bodies, forests, and agricultural land, no systematic vector annotation datasets are currently available.

\subsection{Semantic Segmentation Model}
In pixel-level land cover classification, deep learning methods have become the dominant approach, with numerous researchers proposing segmentation models tailored to remote sensing scenarios. Initially, many methods based on Convolutional Neural Networks (CNNs) were introduced, driving significant advancements in remote sensing image segmentation. For example, Emmanuel Maggiori et al.~\cite{maggiori2017convolutional} proposed an end-to-end pixel classification framework for remote sensing imagery based on Fully Convolutional Networks (FCNs). By employing a two-stage training process and multi-scale module design, this method significantly improved segmentation accuracy, outperforming traditional Support Vector Machine (SVM) classifiers. Jamie Sherrah et al.~\cite{sherrah2016fully} adopted FCNs, achieving more refined land cover boundary segmentation on the Vaihingen and Potsdam datasets. Michele Volpi et al.~\cite{volpi2016dense} introduced a dense semantic segmentation approach based on neural networks, which encodes spatial features and learns deconvolution to restore features to the original resolution, resulting in improved multi-class land cover classification at the pixel level. D. Marmanis et al.~\cite{marmanis2018classification} proposed an end-to-end trainable deep CNN, incorporating semantic boundary detection, which achieved favorable segmentation results on the Vaihingen dataset. Kento Doi et al.~\cite{doi2018effect} addressed class imbalance in training data by introducing a Focal Loss-based approach, providing a more stable optimization path for model training. Building on these advancements, variants of the U-Net model have made significant progress, with the introduction of skip connections leading to substantial improvements in land cover segmentation performance. Andrew Khalel et al.~\cite{khalel2018automatic} enhanced classification performance on the Inria and Massachusetts datasets by concatenating U-Net models. The MACU-Net~\cite{9343296} network extended the U-Net architecture by incorporating multi-scale skip connections and asymmetric convolutions, achieving excellent results on large-scale land cover classification datasets. \added[id=R1,comment={1}]{LANet~\cite{ding2020lanet} enhanced feature representation through the incorporation of both a patch-wise attention module and an attention embedding module, thereby improving semantic segmentation performance in remote sensing images.} With the widespread application of Transformers in computer vision, Transformer-based methods for pixel-level land cover classification have gained increasing attention. The UnetFormer~\cite{wang2022unetformer} network is a hybrid model combining CNN encoders with Transformer decoders, integrating the advantages of both approaches and demonstrating strong classification performance. The BANet~\cite{wang2021transformer} network leverages Transformer structures to model long-range global relationships within images, while convolutional structures extract fine local textures, ultimately fusing the two feature types.

\subsection{Vector Extraction Model}
In the vectorization of land cover features from remote sensing imagery, existing research has primarily focused on two typical man-made objects: buildings and roads. Early methods generally adopted a two-stage strategy of "semantic segmentation + post-processing," where the land cover mask was first obtained using segmentation models, and then geometric rules or heuristic algorithms were applied to generate vector representations. In building vectorization, Zhao et al.~\cite{zhao2018building} initially obtained the building footprints through a segmentation network and subsequently applied the Douglas-Peucker algorithm~\cite{zhang1984fast} to simplify complex polygons, followed by further optimization of the geometric structure based on the minimum description length criterion. Wei et al.~\cite{wei2019toward} proposed a multi-scale fully convolutional network for building segmentation and designed a polygon optimization strategy that includes both coarse fitting and fine correction. In the domain of road vectorization, Cheng et al.~\cite{cheng2017automatic} developed a cascaded convolutional neural network, which simultaneously outputs probability maps for road areas and centerlines, followed by refinement processes to extract the road centerlines. The DeepRoadMapper proposed by Máttyus et al.~\cite{mattyus2017deeproadmapper} utilized a shortest-path connection strategy as a post-processing step to repair fragmented road network structures, enhancing overall topological connectivity.

With the advancement of deep learning and structured modeling capabilities, researchers have recently begun to explore end-to-end land cover vector extraction methods, aiming to address the fragmentation issue between segmentation and post-processing in traditional two-stage approaches. FFL~\cite{girard2021polygonal} method incorporates frame field outputs as one of the model tasks, aligning the predicted frame fields with ground truth contours to provide effective constraints for building regularization, thus improving building footprint detection accuracy. BuildMapper~\cite{wei2023buildmapper} introduced the first end-to-end trainable building vector extraction framework, incorporating a contour evolution module that directly generates regularized building polygons, eliminating the need for complex post-processing. PolyBuilding~\cite{hu2023polybuilding}, based on the Transformer architecture, jointly predicts building vertices and bounding boxes, enabling end-to-end generation of vectorized building outputs with superior contour regularity and precision. HiSup~\cite{xu2023hisup} employs supervisory signals from low-level vertex information to high-level region masks to enhance building polygon extraction and improves prediction accuracy through efficient feature embedding mechanisms. SAMPolyBuild~\cite{wang2024sampolybuild} introduces additional vertex and boundary learning tasks into the SAM decoder, using a mask-guided vertex connection algorithm to predict polygons, achieving higher detection accuracy and better generalization in building vectorization tasks. \added[id=R1,comment={1}]{GCP~\cite{zhang2025global} utilizes a collinearity-aware polygon simplification module to simplify building contours, achieving high-precision end-to-end building contour extraction.} For road targets, Sat2Graph~\cite{he2020sat2graph} predicts node positions and directions in road networks using tensor encoding, enabling direct extraction of road centerlines. RNGDet~\cite{xu2022rngdet} and RNGDet++~\cite{xu2023rngdet++} adopt imitation learning strategies, training the model through iterative path search to gradually reconstruct complete road network structures. SamRoad~\cite{hetang2024segment}, built upon the SAM encoder, incorporates a node connection module to replace traditional complex post-processing, enhancing the quality of road network topology generation. \added[id=R1,comment={1}]{Inspired by Sat2Graph~\cite{he2020sat2graph} and RNGDet~\cite{xu2022rngdet}, GLD-Road~\cite{deng2025gld} employs a hybrid approach that combines global and local road network retrieval, achieving higher efficiency and accuracy in road network extraction.} Additionally, TopDiG~\cite{yang2023topdig} proposes a universal framework for extracting topological structures from remote sensing imagery, capable of simultaneously handling vector representations of both polygonal (e.g., buildings) and linear (e.g., roads) targets, thus broadening the applicability of remote sensing vector extraction methods.

\section{IRSAMap Dataset}
\begin{figure*}[htbp]
  \centering
  \includegraphics[width=\textwidth]{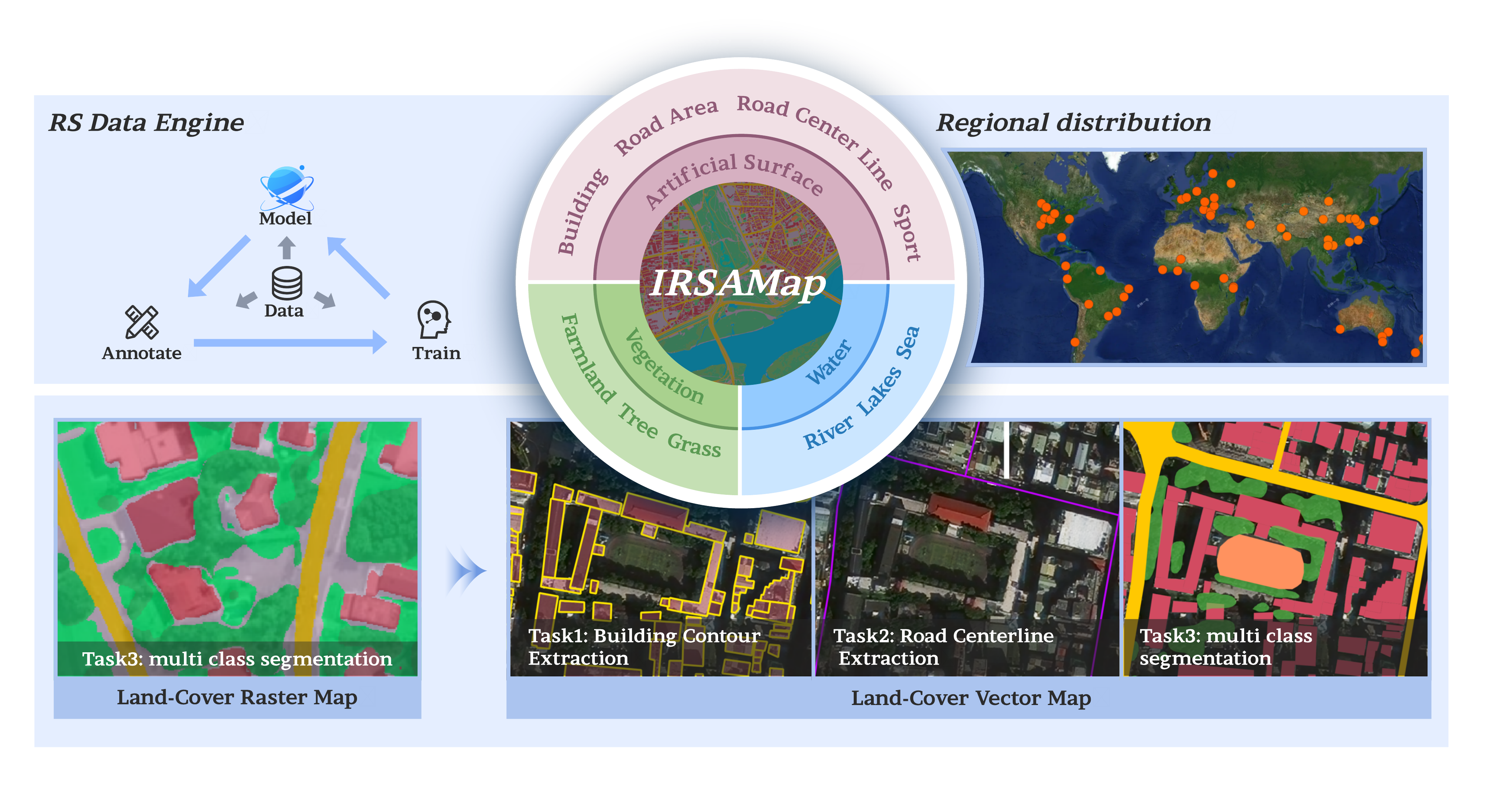} 
  \captionsetup{justification=centering}  
  \caption{Overview of the IRSAMap dataset. The top left shows the dataset's annotation process. The central section displays the annotated classes of the IRSAMap dataset. The regional distribution map (top right) highlights the global coverage of IRSAMap imagery. The bottom left compares traditional raster datasets, which only support semantic segmentation tasks, with IRSAMap, which supports multi-class tasks.}
  \label{fig:fig1}
\end{figure*}

The section provides a comprehensive description of IRSAMap, addressing five key aspects: Data Collection, Annotation Guidelines, Annotation Process, Dataset Statistics and Comparison with Existing Datasets, and Multi-task Support.~\autoref{fig:fig1} presents an overview of the IRSAMap dataset.

\subsection{Data Collection}
During the construction of IRSAMap, remote sensing imagery from various sources and regions was selected to improve the generalization performance of the vectorized mapping model across large-scale areas. The geographical coverage of the imagery spans multiple ecological zones within China as well as various global regions, ensuring data diversity across different areas. Additionally, to address the potential impact of varying image resolutions on the model, a resolution normalization technique is employed, enabling uniform representation of land cover types at a standardized resolution. The following sections provide a detailed description of the data sources, the spatial distribution of the imagery, processing methods, and the dataset partitioning strategy.

\textbf{Data Sources}: The remote sensing imagery utilized in IRSAMap features a spatial resolution of 0.5 meters, sourced from Jilin-1 satellite data, the Google Earth platform, and the Open Cities AI Challenge~\cite{gfdrr2020}. For regions within China, 13 Jilin-1 scenes were acquired. To balance annotation diversity with labeling costs, representative subregions were selected from each scene for fine-grained annotation. This process resulted in a dataset comprising 16 large-scale annotated images covering 16 distinct areas. Notably, three of these samples were extracted from spatially non-overlapping subregions within a single scene, and were specifically designated for the test set to evaluate the model's generalization capability within the same domain.

For most regions outside China, imagery was acquired from the Google Maps platform via the Public Maps Static API. To facilitate manual annotation, imagery at zoom levels 19–20 was selected, ensuring sufficient spatial resolution and visual clarity. For the African region, building footprint annotations provided by the Open Cities AI initiative were used as a baseline. These annotations were manually refined to improve geometric precision, and additional land cover types were vectorized to establish a comprehensive land cover labeling schema. Through the integration of these diverse data sources, IRSAMap offers a high-quality and heterogeneous dataset to support land cover classification and model training.

\textbf{Spatial Distribution of Imagery}: The spatial distribution of IRSAMap imagery within China was guided by the national ecological zoning system, specifically covering nine first-level ecological~\cite{wangzhihua} regions. Representative areas were manually selected within each zone to ensure ecological diversity during data acquisition. For regions outside China, the selection was based on continental divisions across six continents, aiming to maximize geographic representativeness and diversity. \added[id=R1, comment={2}]{The use of different sampling strategies for China and other regions is due to the fact that existing land cover datasets are predominantly concentrated outside of China. However, given China's rich and diverse geographical and topographical features, IRSAMap specifically selected larger regions within China to enhance the performance of the trained model.} IRSAMap integrates remote sensing data from Jilin-1 satellite imagery, the Google Earth platform, and the Open Cities AI challenge. The final dataset comprises approximately 5,000 image tiles, each with a resolution of 1024$\times$1024 pixels, covering 79 representative regions across six continents with a total area of approximately 1,000 km². Specifically, the Jilin-1 imagery covers 16 regions with a total area of 419 km²; data acquired from Google Earth spans 33 regions and 436 km²; and the Open Cities AI data, focused on African urban areas, includes 30 regions covering 155 km². In total, 63 regions outside China are included, accounting for 591 km² of coverage.

\textbf{Resolution Normalization}: IRSAMap involves various natural and anthropogenic land cover elements, each exhibiting significant differences in spatial distribution scales within remote sensing imagery. Directly using the original image resolution may lead to scale inconsistency in multi-element interpretation. Specifically, small-scale features such as roads and buildings are more sensitive to high-resolution imagery, while larger-scale features such as forests and croplands show greater stability with respect to resolution changes. To mitigate the impact of resolution differences on the model training process, IRSAMap applies resolution normalization to both the imagery and vector labels, standardizing the spatial resolution to 0.5 meters. This approach effectively harmonizes scale differences among various land cover types, ensuring that spatial structures and texture features of all elements are accurately represented at a uniform scale. This normalization not only provides consistent input data for the model training process but also establishes a standardized data foundation for subsequent raster-vector consistency modeling, fine-grained boundary extraction, and high-precision map generation.

\textbf{Dataset Partitioning}: The dataset is divided into training, validation, and test sets, with all images standardized to a size of 1024$\times$1024 pixels. The training set consists of 4617 labeled images, covering 44 regions. The validation set includes 340 images across 10 regions, with two regions dedicated to within-domain validation and eight regions allocated to cross-domain validation. The test set contains 912 images, with five regions used for within-domain validation and 20 regions for cross-domain validation.

\subsection{Annotation Guidelines}

\begin{table*}[htbp]
\centering
\caption{Category codes and color representations in IRSAMap Dataset}
\resizebox{0.8\textwidth}{!}{
\begin{tabular}{cccccc}
\hline
\textbf{First-level Category} &  & \textbf{Second-level Category} & &  \\ \hline
\textbf{code} & \textbf{name} & \textbf{code} & \textbf{ref-code} & \textbf{name} &  \textbf{color(RGB)} \\ \hline
  & &  10 & 1 & Farmland & 255,253,145 \\
1 & Vegetation & 11 & 2 & Tree     & 32,216,109 \\
  &            & 12 & 3 & Grass    & 1,252,119 \\
\hline
  & & 21 & 4 & River    & 20,197,246 \\
  2 & Water             & 22 & 5 & Lakes    & 20,185,246 \\
  &            & 23 & 6 & Sea      & 20,197,232 \\
\hline
&
  & 31 & 7 & Building  & 210,75,97 \\
  3 & \begin{tabular}[c]{@{}c@{}}Artificial\\ Surface\end{tabular}   & 32 & 8 & Road\_area & 255,200,1 \\
  &     & 33 & 9 & Road\_line & 192,1,255 \\
  &     & 34 & 10 & Sport     & 255,156,95 \\
\hline
4 & Bareland   & 40 & 11 & Bareland & 204,181,206 \\
\hline
\end{tabular}
}
\label{tab:classidtab}
\end{table*}

Following the principle of "what you see is what you get," vector labels are constructed for the collected imagery data by stratifying according to different land cover types. For instance, anthropogenic features such as roads and buildings are vectorized through detailed instance-level annotations, while natural features like vegetation are annotated with fine boundary delineation. This stratified labeling approach allows IRSAMap to not only support traditional image segmentation methods for land cover tasks but also facilitates multi-task research, enabling researchers to flexibly combine different label layers for diverse analyses. As shown in~\autoref{tab:classidtab}, the IRSAMap dataset includes four first-level categories (Vegetation, Water Body, Artificial Surface, Bareland) and 11 second-level categories (Farmland, Tree, Grass, River, Lakes, Sea, Building, Road Area, Road Centerline, Sport, Bareland). Compared to previous multi-feature land cover datasets~\cite{wang2loveda, xia2023openearthmap}, IRSAMap provides a more granular classification of data categories.

\textbf{Vegetation}: The Vegetation category is further subdivided into three subcategories: Farmland, Tree, and Grass. Notably, the annotation of trees within IRSAMap is particularly detailed. The dataset includes not only large forested areas but also individual trees within urban blocks, along road medians, and in extensive grasslands. These trees, which are often merged with other land cover types or overlooked in other multi-feature land cover datasets, are distinctly identified in IRSAMap. In areas where trees and grass are difficult to differentiate, particularly in shadowed regions, annotations are carefully delineated based on the continuity of the grass cover to ensure accurate differentiation.

\textbf{Water Body}: The Water Body category is divided into three subcategories: River, Lakes, and Sea. In the annotation of the boundaries of Lakes and Sea, care was taken to trace the visible natural boundaries rather than smoothing based on human interpretation. Additionally, urban channels and narrow rivers are included in the River category. Given the potential for confusion between channels and other land cover types such as roads, particular attention was paid to the connectivity of the channels during annotation to ensure accurate classification.

\textbf{Artificial Surface}: The Artificial Surface category is divided into four subcategories: Building, Road Area, Road Line, and Sport. For the annotation of buildings, individual buildings are traced strictly according to their actual outlines. In dense urban areas, individual buildings are distinguished based on roof color variations, while large buildings are differentiated based on their boundary contours, with careful alignment of corner points. In the road annotation process, particular attention is paid to the connectivity of roads. For Road Line annotations, OSM data is used as a reference, and each road is adjusted to ensure accurate centering of the road centerline. Road Area annotations are derived from the road centerline, with buffering applied based on measured road widths, followed by secondary corrections for areas where road width changes or at intersections to ensure precise boundary alignment. Outdoor sports fields, such as synthetic tracks and basketball courts, are uniformly annotated as Sport, based on their outer boundaries.
 
\textbf{Bareland}: During the annotation process, all areas not classified under the three aforementioned categories are labeled as Bareland.

\added[id=R2,comment={3}]{\textbf{Annotation Challenges}:In urban areas, the presence of multi-lane roads and interchanges poses significant challenges for road network annotation. Additionally, due to the angle differences between satellites and high-rise buildings, there is often a misalignment of features such as building rooftops and ground structures, which further complicates the annotation process for urban buildings. The boundary between forests and grasslands in urban green spaces is also ambiguous, adding to the complexity of the labeling task. In rural areas, the lack of OSM data in certain regions, coupled with tree occlusion, significantly hampers the manual annotation of road networks. Furthermore, the intricate distribution of farmland, forest, and grassland, along with blurred boundaries between these land cover types, leads to confusion and increases the difficulty of the annotation process in rural regions.}

\subsection{Annotation Procedure}
Following the process outlined in SAM~\cite{kirillov2023segment}, IRSAMap employs a three-phase iterative annotation workflow: "manual pre-annotation + AI-based training and inference + manual review and correction." The RS Data Engine was developed to facilitate large-scale human-AI collaborative annotation. With the assistance of the RS Data Engine, the annotation of 1000 square kilometers of remote sensing imagery was completed with limited human resources (6 staff members, contributing a total of approximately 1500 hours). This effort resulted in the creation of a land cover vector mapping dataset, IRSAMap, which includes 1.8 million masks. The specific process is outlined as follows:
\begin{figure}[htbp]
  \centering
  \includegraphics[width=\columnwidth]{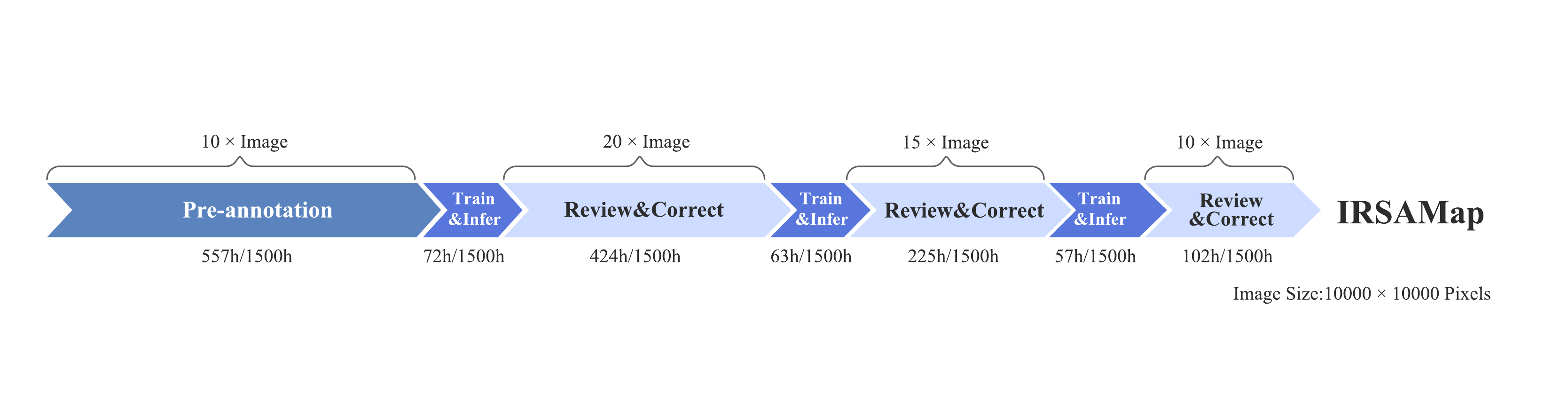} 
  \captionsetup{justification=centering}  
  \caption{\textcolor{white}{Time cost per phase and overall workflow in the IRSAMap annotation procedure.}}
  \label{fig:wordflow}
\end{figure}

\textbf{Manual Pre-Annotation}: Ten large image tiles, each approximately 10,000$\times$10,000 pixels, were manually annotated with fine detail. The annotation time per tile varied by class due to differences in complexity: vegetation required approximately 26 hours, buildings 23 hours, roads 7 hours, sports facilities 0.5 hours, and water bodies 4 hours. Vegetation annotation was particularly time-consuming due to its fragmented and heterogeneous nature. Building annotation involved meticulous delineation of individual structures, especially in dense urban areas; the most labor-intensive case involved a single tile containing approximately 40,000 individual buildings. Road annotation was comparatively more efficient, aided by reference to OSM data.

\textbf{AI-based training and inference}: The annotated data undergo preprocessing to generate image tiles suitable for model training. During this phase, a multi-class semantic segmentation model is trained for water bodies, vegetation, buildings, and sports facilities. A separate model is trained for road networks due to their higher demands on global topological consistency. Upon completion of model training, the resulting models are used to perform inference on the remaining unlabeled image regions.

\textbf{Manual Review and Correction}: The segmentation model outputs were first subjected to vector topology simplification. Annotators then reviewed these vectorized results, correcting or removing erroneous segments as needed. The corrected annotations were subsequently incorporated into retraining and inference cycles. \added[id=R2, comment={1}]{As illustrated in~\autoref{fig:wordflow}, three iterations of this process were conducted to complete the full annotation of the IRSAMap dataset.} During the final correction round, the annotation time per image tile was reduced to approximately one-sixth of the time required for manual annotation from scratch.

\subsection{Dataset Statistics and Comparison with Existing Datasets}
\begin{table*}[htbp]
\centering
\caption{Dataset statistics for training, validation, and testing sets}
\resizebox{0.8\textwidth}{!}{
\begin{tabular}{cccccccc}
\hline
\textbf{Category} & \multicolumn{2}{c}{\textbf{Train}} & \multicolumn{2}{c}{\textbf{Validation}} & \multicolumn{2}{c}{\textbf{Test}} \\
\cline{2-7}
& \textbf{Count (units)} & \textbf{Area (km\textsuperscript{2})} & \textbf{Count (units)} & \textbf{Area (km\textsuperscript{2})} & \textbf{Count (units)} & \textbf{Area (km\textsuperscript{2})} \\
\hline
Bareland    & 758339  & 303.27 & 66766  & 16.95   & 146180 & 58.73 \\
Farmland    & 7963    & 117.50  & 512    & 2.02    & 963    & 15.58  \\
Tree        & 429879  & 237.92  & 19192  & 18.14   & 66681  & 61.75 \\
Grass       & 276159  & 303.42 & 28471  & 14.69   & 63633  & 43.61 \\
River       & 3454    & 32.28  & 162    & 2.21    & 604    & 18.43  \\
Lakes       & 4855    & 12.74   & 263    & 3.57   & 2026   & 7.40  \\
Sea         & 378     & 18.65   & 226    & 20.63   & 31     & 9.86  \\
Building    & 606316  & 123.64  & 22814  & 6.28   & 82984  & 33.42 \\
Road area   & 35049   & 59.27  & 3315   & 4.44   & 5501   & 14.98  \\
Sport       & 816     & 1.63    & 74     & 0.19     & 214    & 0.48   \\
\hline
\end{tabular}
}
\label{tab:dataset_stats}
\end{table*}

\begin{figure}[htbp]
  \centering
  \includegraphics[width=\columnwidth]{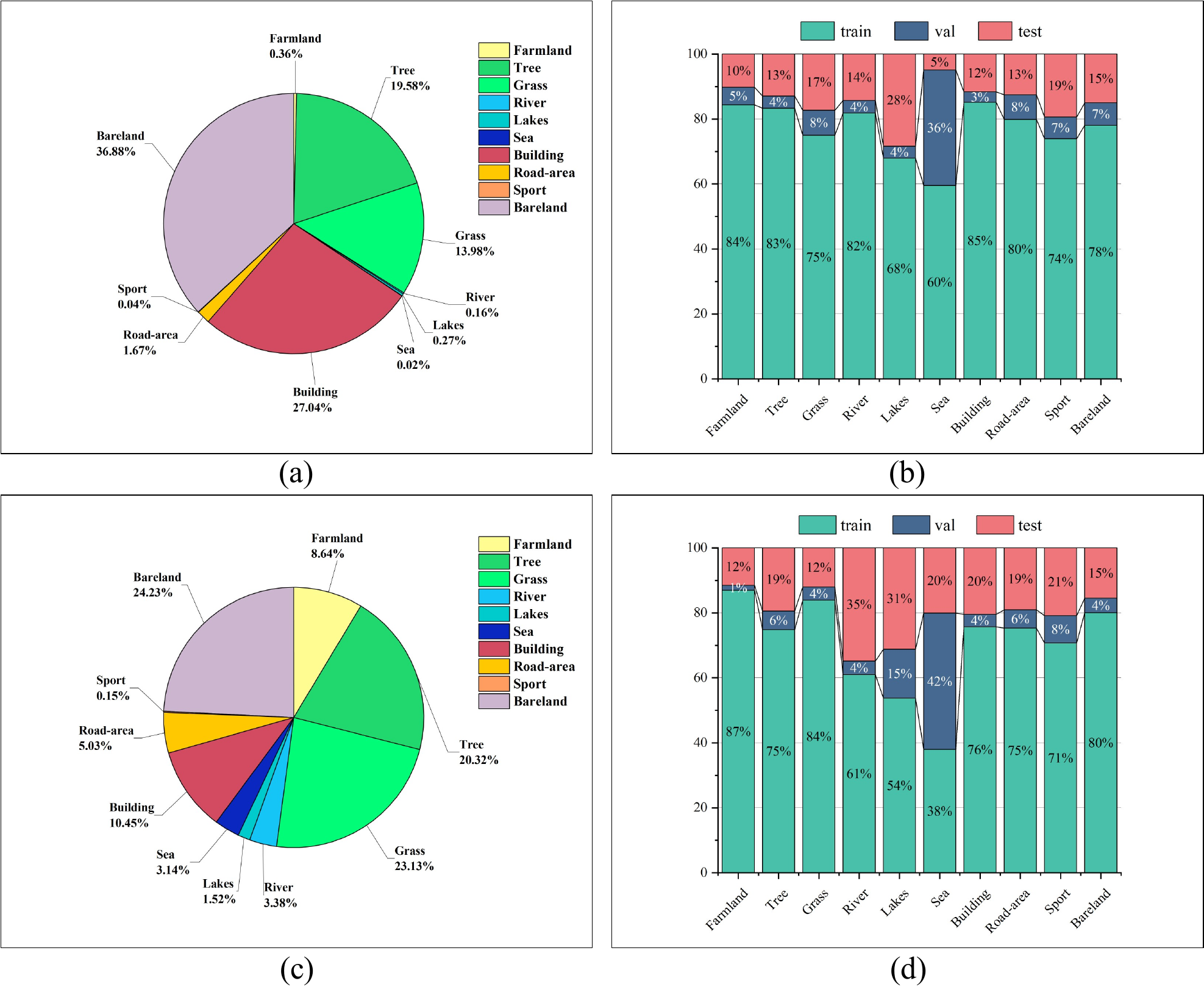} 
  \captionsetup{justification=centering}  
  \caption{Statistical analysis chart of object categories in the IRSAMap dataset.}
  \label{fig:any}
\end{figure}

The IRSAMap dataset is divided into three parts: training set, validation set, and test set. A systematic statistical analysis of the instance count and their corresponding areas for each category has been conducted. As shown in~\autoref{tab:dataset_stats}, the instance count for each semantic category is detailed.~\autoref{fig:any}(a) shows that the Tree, Grass, and Building categories have a relatively large number of instances, indicating that IRSAMap contains fine-grained instance-level annotations.~\autoref{fig:any}(b) illustrates the proportion of each category in different splits, which is generally balanced.~\autoref{fig:any}(c) indicates that the Sport class, due to its inherently small quantity, has a lower area proportion, which increases the detection difficulty.~\autoref{fig:any}(d) reveals that the proportions of most categories remain consistent across the three subsets, while the Sea class has a higher proportion in the validation set than in the training set. This is primarily because when it appears, it often occupies the majority of the area in only a few images.

Compared to the current high-resolution land cover dataset, OpenEarthMap, IRSAMap extends the capability of rasterized land cover representation in terms of annotation standards and land cover category definitions. Building upon the annotation systems of LoveDA and OpenEarthMap, IRSAMap offers greater flexibility in task combinations, supporting multi-task and multi-granularity spatial object modeling. In terms of data scale, IRSAMap covers an area of 1000 km², slightly surpassing OpenEarthMap's 970 km². Additionally, IRSAMap provides more detailed semantic annotations, enabling support for more complex downstream interpretation tasks.

In terms of road centerline annotations, IRSAMap demonstrates superior geographic diversity and a significantly larger annotation scale compared to the City-Scale dataset. While City-Scale is limited to road scenes from 20 cities within the United States, comprising 180 images at a resolution of 2048$\times$2048 pixels, IRSAMap includes approximately 5,000 images of 1024$\times$1024 pixels, covering over 70 regions across six continents. Furthermore, IRSAMap employs imagery with a spatial resolution of 0.5 meters, offering finer detail than City-Scale's 1-meter resolution. The total annotated area in IRSAMap is approximately 1.3 times that of City-Scale. To enhance topological accuracy, all road centerlines in regions with inaccurate OSM data were manually refined by annotators. In the IRSAMap dataset, the annotated lengths of road centerlines amount to 8532.60 km for the training set, 684.51 km for the validation set, and 2003.89 km for the test set.

In terms of vector annotation for individual buildings, IRSAMap offers improved instance-level delineation compared to the existing WHU-Mix(vector) dataset. Although WHU-Mix(vector) integrates multiple data sources and applies boundary refinement, it exhibits limitations in building instance separation, with some images still containing merged or contiguous building annotations. Statistical analysis shows that each 512$\times$512 image in WHU-Mix(vector) contains an average of 12 buildings, with a maximum of 230. In contrast, IRSAMap provides fine-grained instance-level annotations across diverse urban and rural environments, with an average of 32 buildings per 512$\times$512 image and a maximum of 685. In total, IRSAMap includes approximately 730,000 individual building instances, offering a significantly enhanced resource for structured building segmentation tasks.

\subsection{Multi-Task Learning Support}
\begin{figure}[htbp]
  \centering
  \includegraphics[width=\columnwidth]{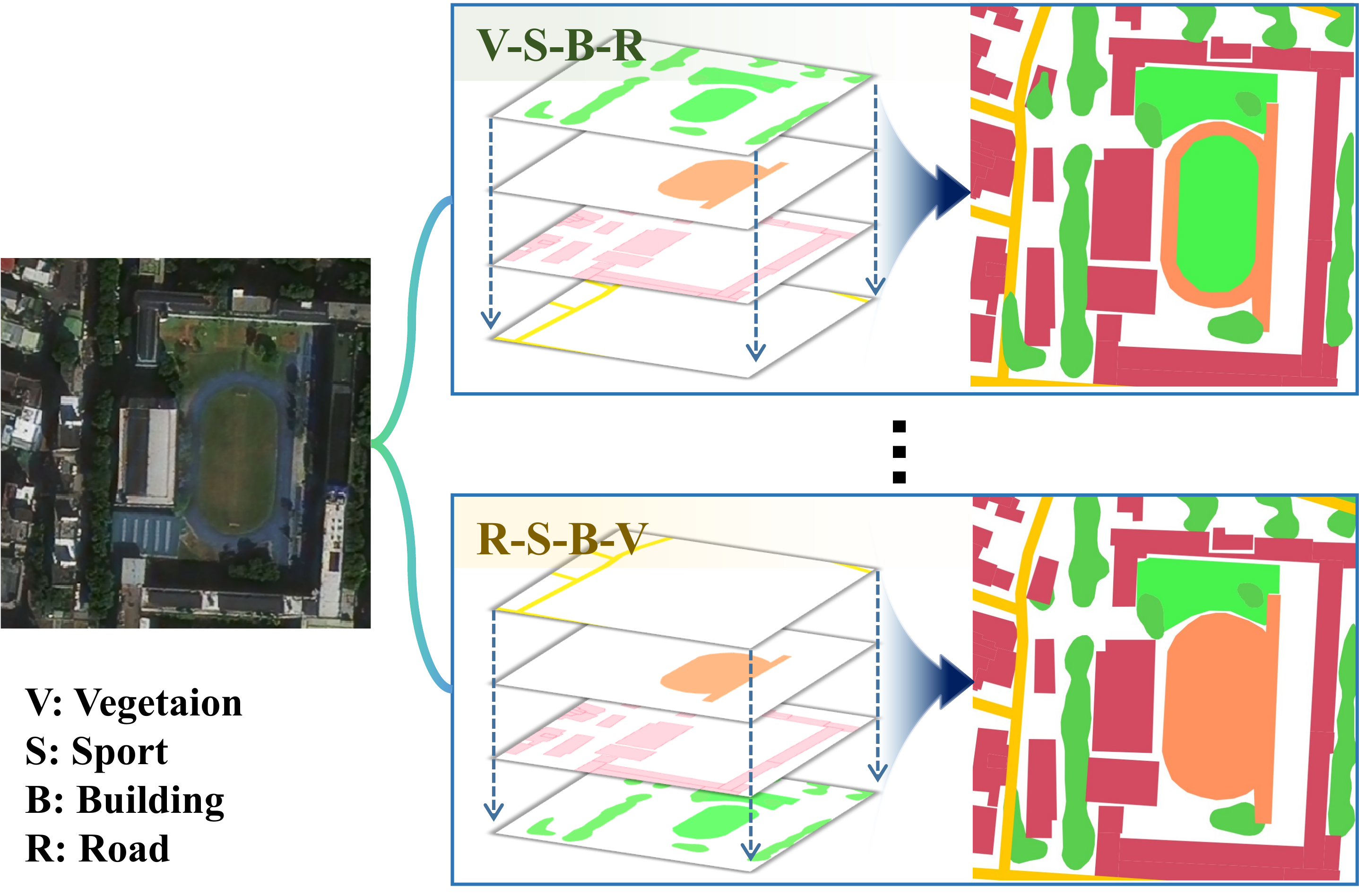} 
  \captionsetup{justification=centering}  
  \caption{Schematic Diagram of Flexible Layer Composition.}
  \label{fig:fig2}
\end{figure}

IRSAMap is the first dataset to provide multi-feature vector cover annotations, enabling the integration of natural land cover segmentation tasks and object-level vector extraction tasks for anthropogenic features. This approach facilitates the construction of high-precision land cover vector maps. IRSAMap supports three distinct types of geographic information extraction tasks: land cover segmentation, road centerline and boundary extraction, and building outline extraction.
For the land cover segmentation task, IRSAMap provides multi-level vectorized land cover annotations. Through vector-to-raster conversion and multi-feature raster overlay, IRSAMap not only supports multi-class land cover segmentation tasks commonly found in datasets such as LoveDA~\cite{wang2loveda} and OpenEarthMap~\cite{xia2023openearthmap}, but also offers enhanced flexibility in task design. Users can adjust the layering sequence according to specific task requirements. For example, in a road connectivity mapping scenario (as shown in~\autoref{fig:fig2}), 
users can prioritize the overlay of the road layer to ensure network connectivity, whereas in a vegetation coverage analysis, the vegetation layer can be given higher priority. This feature overcomes the limitations of traditional raster-based datasets in handling feature conflicts and spatial relationship representation, providing greater flexibility for task customization.

\begin{figure}[htbp]
  \centering
  \includegraphics[width=\columnwidth]{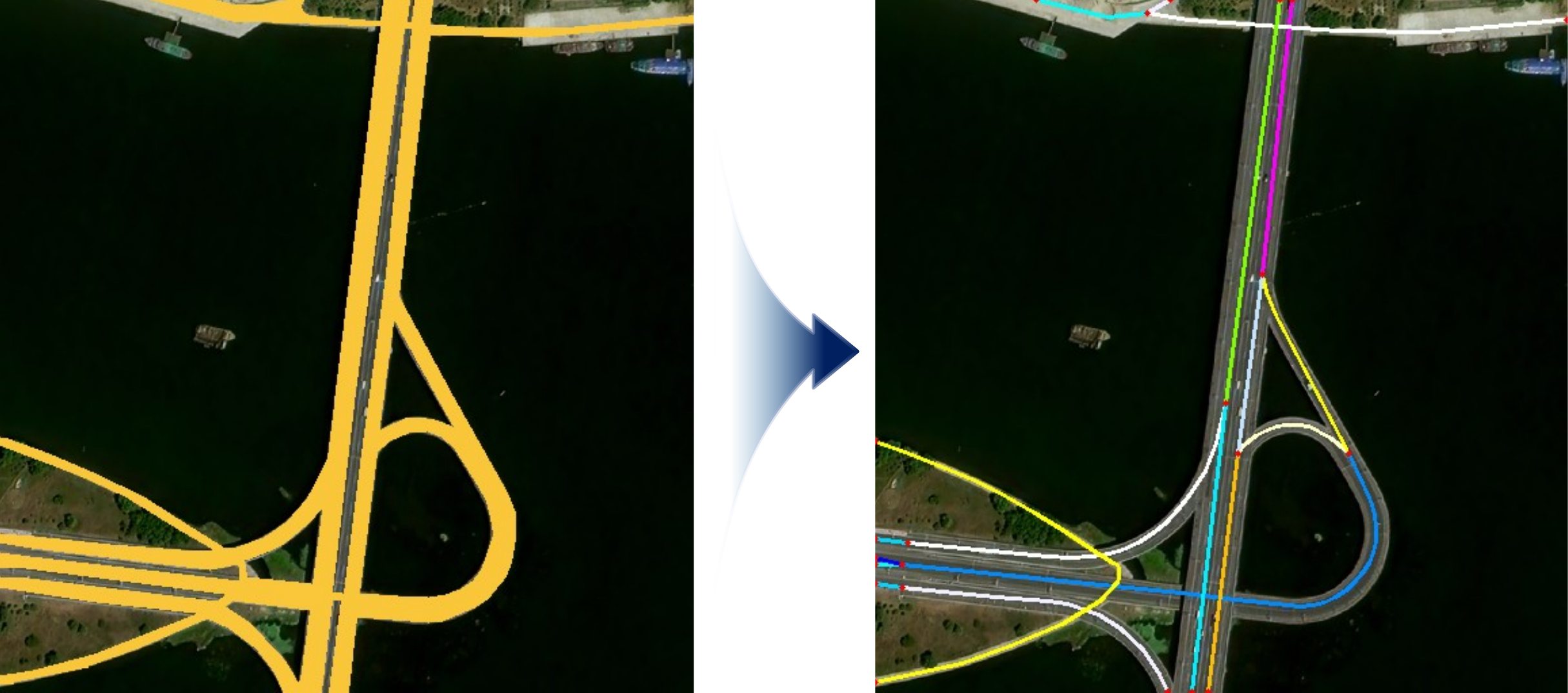} 
  \captionsetup{justification=centering}  
  \caption{Comparison of Raster and Vector Road Network Annotation.}
  \label{fig:fig3}
\end{figure}

For road centerline and boundary extraction tasks, IRSAMap provides vector annotations, including both road boundary and centerline, supporting more refined road extraction models that go beyond traditional semantic segmentation. These models incorporate point, line, and polygon-based modeling methods~\cite{he2020sat2graph, xu2022rngdet, tan2020vecroad, hu2023polyroad, zao2023topology, hetang2024segment, li2024roadcorrector, yang2024roaddet, yin2024towards}. \added[id=R1,comment={4}]{Due to the overhead perspective of remote sensing imagery, the rasterized road network labels struggle to distinguish between interchanges and intersections. However, vectorized labels provide a clearer representation of the topological relationships between road segments, effectively alleviating road ambiguity. As shown in~\autoref{fig:fig3}, the visualization of vectorized labels distinctly separates each road segment, using different colors to represent them, with red dots marking the endpoints or intersections. Notably, no intersection points are generated on the roads of interchanges.}

\begin{figure}[htbp]
  \centering
  \includegraphics[width=\columnwidth]{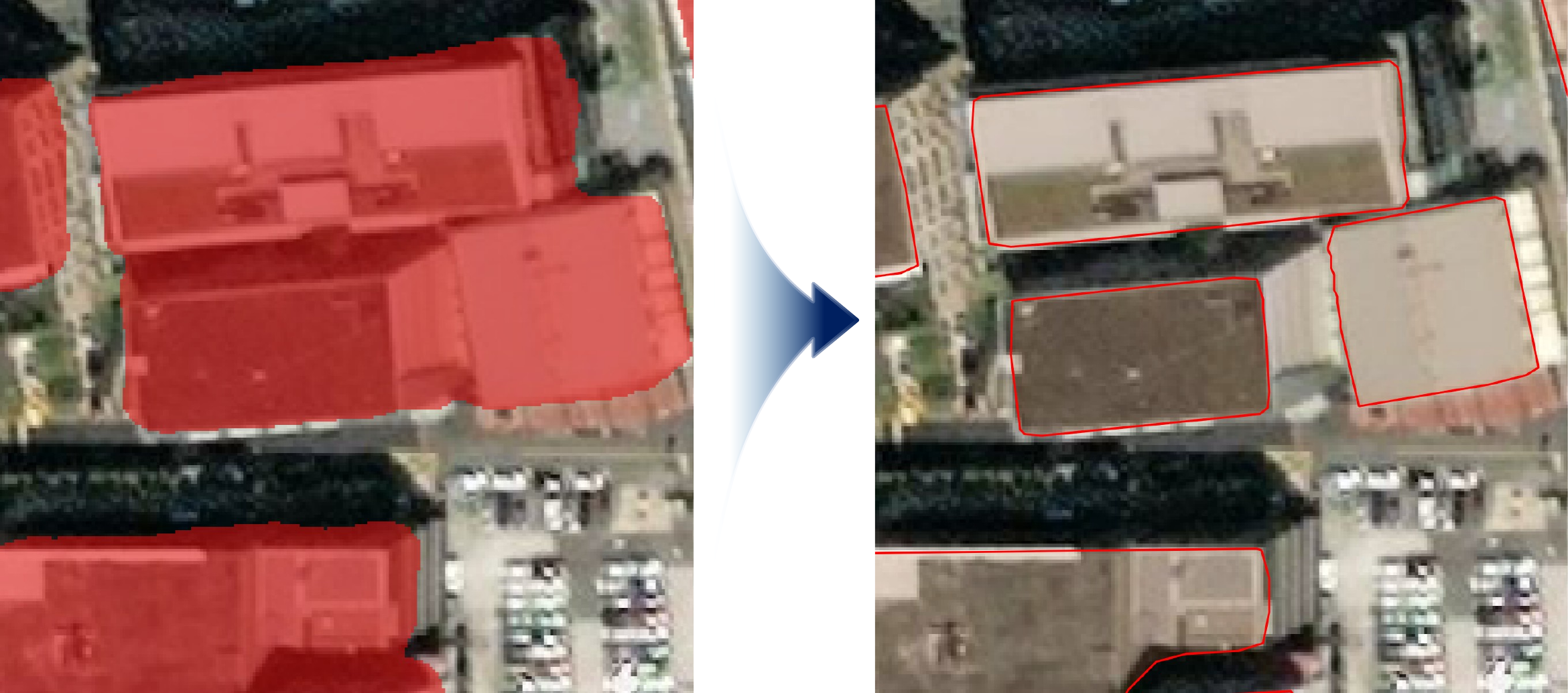} 
  \captionsetup{justification=centering}  
  \caption{Comparison of Raster and Vector Building Annotations.}
  \label{fig:fig4}
\end{figure}

For building instance detection tasks, IRSAMap provides over 700,000 high-quality building instance contours, covering building groups of varying scales, densities, and forms. The dataset is particularly suited for studies in high-density urban areas, where a single 512$\times$512 pixel remote sensing image may contain more than 300 building instances, reflecting the dense building characteristics typical of developed urban environments. This makes IRSAMap one of the largest and highest-quality publicly available building instance vector datasets. The building instance boundaries in IRSAMap are finely delineated from high-resolution remote sensing imagery, accurately preserving the true edge features of buildings, including irregular contours, ancillary structures (such as roof components and stairwells), and gaps between buildings. The precise representation of these features surpasses traditional coarse or rectangular-fitting annotations, improving geometric accuracy in instance segmentation and reducing errors due to building overlap, as shown in~\autoref{fig:fig4}.

\section{Baseline Experience}
In this section, we evaluate the performance of IRSAMap on three representative tasks using state-of-the-art model approaches. \added[id=R2, comment={2}]{All experiments in this section were conducted on a machine equipped with 512GB of memory and four 3090 GPUs, with the models trained and inferred using the PyTorch framework. To ensure the rigor of the experiments, the data augmentation methods, hyperparameter settings, and optimizers for the eleven experimental models (UPerNet~\cite{xiao2018unified}, UnetFormer~\cite{wang2022unetformer}, FT-UnetFormer~\cite{wang2022unetformer}, RNGDet~\cite{xu2022rngdet}, RNGDet++~\cite{xu2023rngdet++}, Sam-Road~\cite{hetang2024segment}, GLD-Road~\cite{deng2025gld}, FFL~\cite{girard2021polygonal}, HiSup~\cite{xu2023hisup}, SAMPolyBuild~\cite{wang2024sampolybuild}, and GCP~\cite{zhang2025global}) were kept consistent with those specified in the respective papers and open-source codebases. However, it is worth noting that due to limited hardware resources, the training batch size for GCP was set to 2, and distributed training was performed across the four 3090 GPUs.}

\subsection{Land cover Segmentation}

\begin{table*}[htbp]
\centering
\caption{Results of the Land Cover Task on the IRSAMap Dataset.}
\resizebox{\textwidth}{!}{
\begin{tabular}{ccccccccccccc}
\hline
\textbf{Methods} & \textbf{Backbone} & \textbf{Infer Time} & \multicolumn{8}{c}{\textbf{IoU(\%)}} & \textbf{mIoU(\%)} \\ \hline
             & & &  \textbf{Farmland} & \textbf{Tree} & \textbf{Grass} & \textbf{Water body} & \textbf{Building} & \textbf{Road} & \textbf{Sport}  & \textbf{Bareland} &  \\ \hline
UPerNet\textcolor{white}{~\cite{xiao2018unified}} & Convxt-T & 90.54 & 58.63 & 83.86 & 64.14 & 95.01 & 75.06 & 49.91 & 52.86 & 64.25 & 67.97 \\ 
       & Convxt-S & 106.84 & 58.42 & 84.18 & 63.47 & 95.87 & 76.26 & 51.81 & 52.81 & 64.70 & 68.55 \\ 
       & Convxt-B & 128.14 & 59.72 & 85.07 & 65.35 & 95.77 & 76.30 & 52.10 & 55.22 & 65.00 & 69.26 \\ 
       & Swin-T   & 94.34 & 49.86 & 83.67 & 63.22 & 96.04 & 75.92 & 51.45 & 53.22 & 64.30 & 67.17 \\ 
       & Swin-S   & 111.12 & 58.78 & 84.53 & 64.84 & 96.05 & 76.01 & 52.58 & 55.22 & 64.34 & 69.25 \\ 
       & Swin-B   & 130.34 & 62.07 & 85.07 & 65.79 & 95.51 & 75.88 & 52.27 & 54.41 & 64.72 & 69.25 \\ \hline
UnetFormer\textcolor{white}{~\cite{wang2022unetformer}} & Resnet-18 & 19.09 & 53.26 & 81.23 & 61.52 & 94.58 & 72.97 & 47.66 & 44.98 & 61.19 & 64.67 \\ 
          & Resnetxt-101 & 57.09 & 52.26 & 83.27 & 62.85 & 93.77 & 75.32 & 51.05 & 52.70 & 63.13 & 66.79 \\ \hline
FT-UNetFormer\textcolor{white}{~\cite{wang2022unetformer}}  & Swin-B & 94.42 & 59.79 & 84.00 & 64.47 & 96.10 & 76.43 & 52.92 & 54.97 & 64.38 & 69.36 \\ \hline
\end{tabular}
}
\label{tab:landcovertab}
\end{table*}

\begin{figure}[htbp]
  \centering
  \includegraphics[width=\columnwidth]{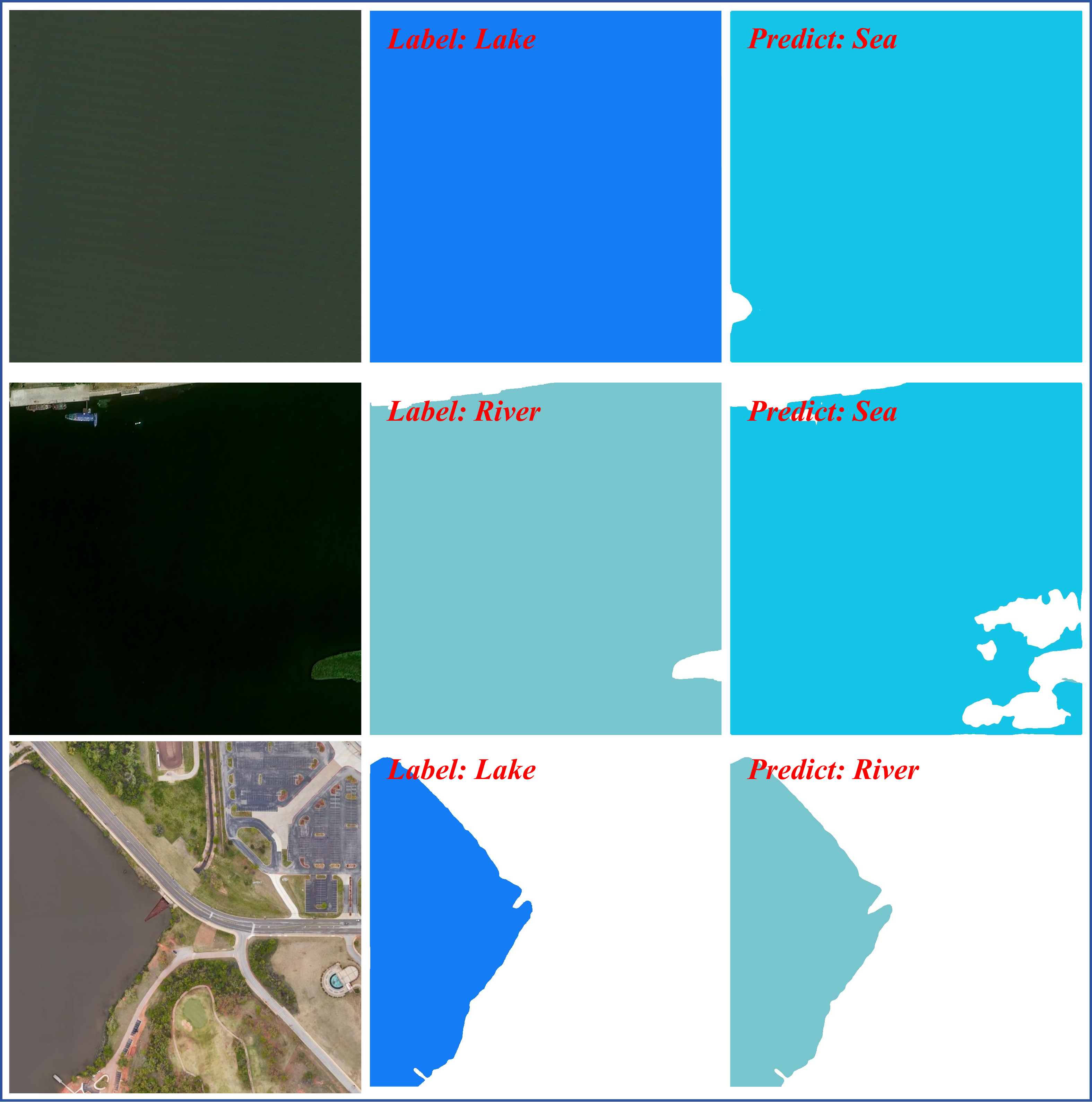} 
  \captionsetup{justification=centering}  
  \caption{Misclassification Examples for Sea, Lake, and River Classes.}
  \label{fig:water_anl}
\end{figure}

\begin{figure*}[htbp]
  \centering
  \includegraphics[width=\textwidth]{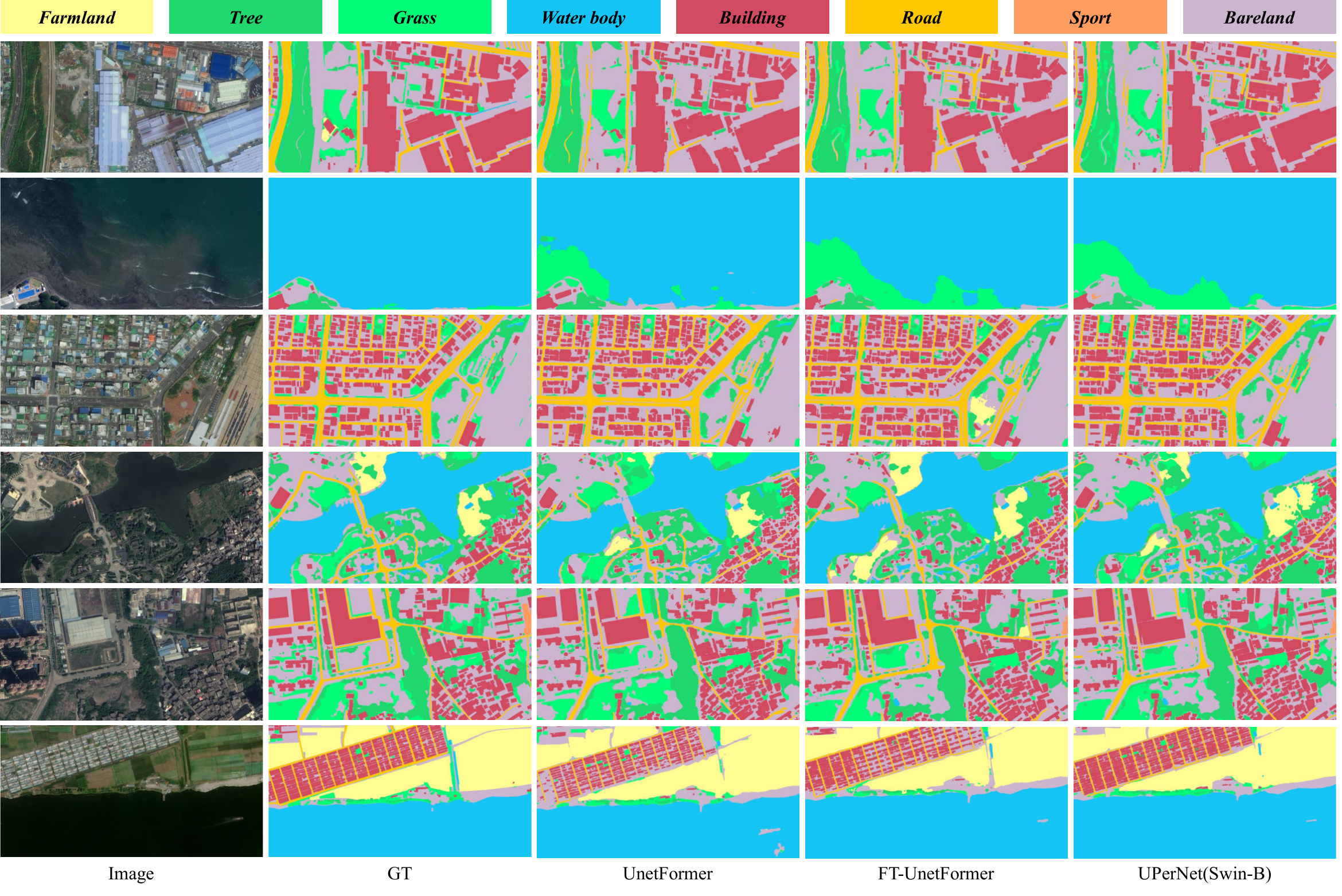} 
  \captionsetup{justification=centering}  
  \caption{Visualization Comparison of Model Results for Land Cover Task.}
  \label{fig:fig5}
\end{figure*}

Benchmarks and Evalution: In this section, we selected UPerNet, UNetFormer, and FT-UNetFormer as representative segmentation models. These models were trained on the IRSAMap dataset to evaluate their performance in land cover classification tasks. To provide a comprehensive assessment of segmentation accuracy, we adopted the mean Intersection over Union (mIoU) as the primary evaluation metric. The mIoU is computed as follows:
\begin{equation}
  \text{mIoU} = \frac{1}{N} \sum_{i=1}^{N} \frac{|A_i \cap B_i|}{|A_i \cup B_i|}
\end{equation}

In the above equation, $N$ denotes the total number of semantic classes. 
$A_i$ represents the set of pixels labeled as class $i$ in the ground truth, 
while $B_i$ denotes the set of pixels predicted as class $i$ by the model. 
The term $|A_i \cap B_i|$ indicates the number of pixels correctly predicted as class $i$, 
i.e., the cardinality of the intersection between ground truth and prediction. 
The term $|A_i \cup B_i|$ denotes the number of pixels belonging to either the ground truth or predicted class $i$, i.e., the cardinality of their union.

Results Analysis: As shown in~\autoref{tab:landcovertab}, the UPerNet model with Swin-B as the backbone achieved the highest mIoU, albeit with the longest inference time. This indicates that increased model complexity contributes to improved segmentation accuracy. \added[id=R1, comment={6}]{Additionally, for~\autoref{tab:landcovertab}, it is important to note that sea, river, and lake were not treated as separate classes. When predicted individually, these subclasses exhibited highly unstable IoU performance, with peak values of 22.96\%, 41.11\%, and 13.90\%, respectively, while in some epochs, the IoU for the sea category dropped below 2\%. This instability adversely affected the overall accuracy evaluation. To mitigate this issue, they were merged into a unified Water Body category for both training and prediction. As illustrated in~\autoref{fig:water_anl}, this variability is largely attributed to the difficulty of distinguishing among these subclasses in small-scale image tiles, where spectral and spatial similarities often lead to misclassification.} Upon further examination of class-wise performance, the extraction of Farmland, Road, and Sport categories was found to be more challenging, as reflected by their relatively lower accuracy. Specifically, Farmland suffers from low accuracy due to its similarity to Grass, while the segmentation precision of the Road class is limited by spectral differences among various road types, such as urban highways, rural roads, and national roads, as well as occlusion from trees. For the Sport class, poor detection accuracy is attributed to a small sample size and limited coverage in the remote sensing images. In the visualization results shown in~\autoref{fig:fig5}, we observe frequent confusion between the Tree and Grass categories across the three segmentation models. Although waterbody detection achieves high accuracy,~\autoref{fig:fig5}(second row) illustrates that, due to the clarity of the water, submerged algae are misidentified as Grass, leading to incorrect boundary identification of the waterbody. Furthermore, as shown in the last row of~\autoref{fig:fig5}, the Building class tends to exhibit a "clumping" effect during segmentation. To mitigate this, a vectorized contour detection method was employed to effectively reduce the occurrence of this phenomenon in the detection of individual buildings.


\subsection{Road Centerline Extraction}
Benchmarks and Evalution: For the road centerline extraction task, we selected several state-of-the-art models that have demonstrated superior performance in road network detection, including RNGDet, RNGDet++, SamRoad, and GLD-Road. These models were retrained on the IRSAMap dataset. Inference on the test set was performed to evaluate the connectivity of the road network. To comprehensively assess the accuracy of the predictions, we employed two metrics: APLS~\cite{van2018spacenet} and TOPO~\cite{biagioni2012inferring}. Specifically, the APLS metric evaluates the global topological connectivity of the predicted road network by measuring the shortest path difference between any two vertices in the ground truth and predicted road networks. This metric reflects the degree to which the global topological structure of the predicted road network matches the ground truth. The TOPO metric, on the other hand, focuses on local connectivity. It assesses the match between the predicted and ground truth road nodes within local regions by comparing the reachable nodes in those regions. These two metrics provide complementary evaluations of road network connectivity, considering both global and local perspectives.
\begin{table}[htbp]
\centering
\caption{Results of the Road Network Extraction Task on the IRSAMap Dataset.}
\resizebox{\columnwidth}{!}{
\begin{tabular}{ccccc}
\hline
\textbf{Methods} & \textbf{APLS} & \textbf{TOPO-P} & \textbf{TOPO-R} & \textbf{TOPO-F1} \\ \hline
RNGDet\textcolor{white}{~\cite{xu2022rngdet}}   & 55.81 & 89.68 & 60.95 & 72.57 \\ 
RNGDet++\textcolor{white}{~\cite{xu2023rngdet++}} & 51.55 & 85.92 & 61.82 & 71.91 \\ 
SamRoad\textcolor{white}{~\cite{hetang2024segment}}  & 58.04 & 94.19 & 58.8  & 72.40 \\ 
GLD-Road\textcolor{white}{~\cite{deng2025gld}} & 57.23 & 85.15 & 68.12 & 75.69 \\ \hline
\end{tabular}
}
\label{tab:roadnetwork_results}
\end{table}

\begin{figure*}[htbp]
  \centering
  \includegraphics[width=\textwidth]{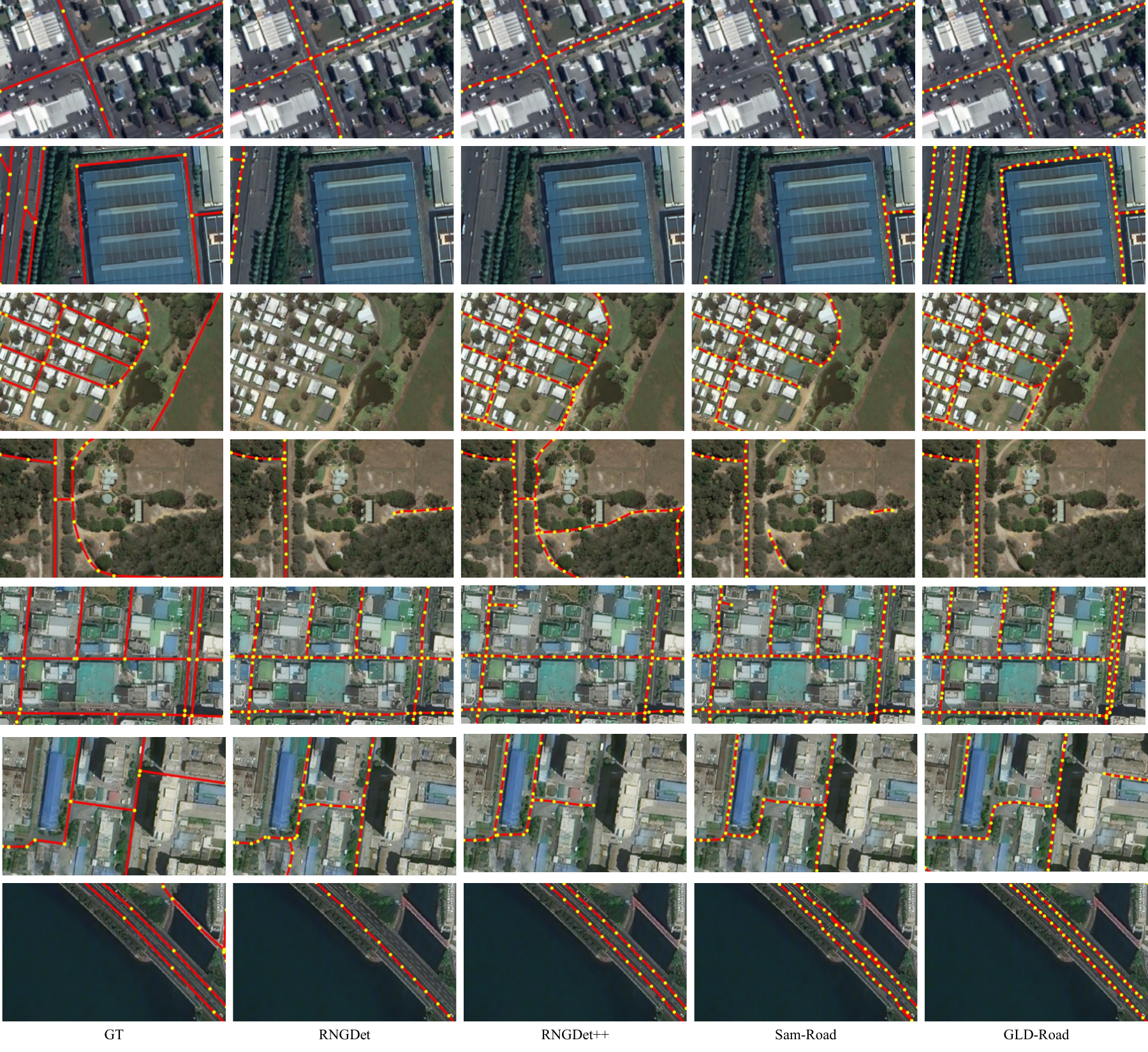} 
  \captionsetup{justification=centering}  
  \caption{Visualization Comparison of Model Results for Road Network Extraction Task.}
  \label{fig:fig6}
\end{figure*}

Results Analysis: Quantitative evaluation results for the four models on the IRSAMap dataset are shown in~\autoref{tab:roadnetwork_results}. In terms of global connectivity as measured by the APLS metric, SamRoad, leveraging the SAM encoder in combination with the Vit-B architecture, demonstrated superior feature extraction capabilities, achieving the highest accuracy of 58.04. In contrast, with regard to the TOPO metric, GLD-Road, benefiting from its local road network completion module, exhibited higher local topological precision, surpassing the other methods on the TOPO-F1 score. On the other hand, both RNGDet and RNGDet++ showed relatively stable performance across the APLS and TOPO metrics. The primary reason for this lies in the iterative search strategy employed by these models, which led to significant road loss, thereby impacting their accuracy. In the road network extraction task, the visual results shown in~\autoref{fig:fig6} reveal that GLD-Road outperforms the other three models in detecting dual-lane roads. RNGDet and RNGDet++ occasionally miss entire roads, while SamRoad tends to lose road segments and often detects straight roads as curved, which does not align with the inherent characteristics of roads. Although RNGDet and RNGDet++ leverage the advantages of iterative search methods, resulting in occasional road loss, they tend to generate road networks with fewer disconnected segments. In contrast, SamRoad, benefitting from the powerful feature extraction capabilities of SAM and the node connectivity of the Topology Decoder, achieves a better balance between road network detection accuracy and connectivity, thus securing a significant advantage in the APLS metric.

\subsection{Building Polygon Extraction}
\begin{figure*}[htbp]
  \centering
  \includegraphics[width=\textwidth]{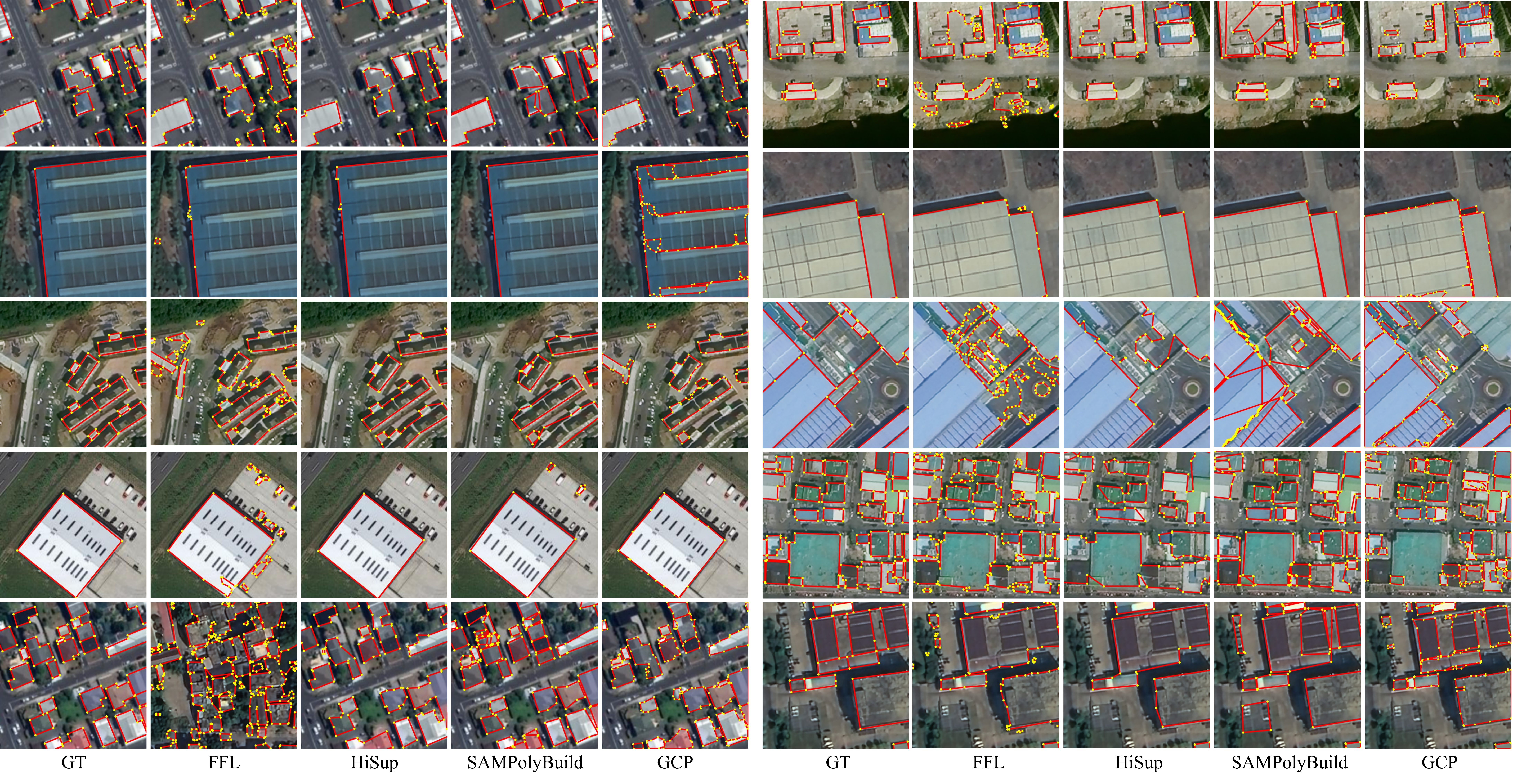} 
  \captionsetup{justification=centering}  
  \caption{Visualization Comparison of Model Results for Building Outline Extraction Task.}
  \label{fig:fig7}
\end{figure*}

Benchmarks and Evalution: AP and AR are key metrics for evaluating the precision and recall of the model in building polygon instance segmentation. Specifically, AP\_poly represents the precision of the model's polygon outputs after converting them into masks, while AR\_poly denotes the recall calculated from the mask conversion of the predicted building polygons. In contrast, AP\_mask evaluates the precision based on the masks directly output by the model. Additionally, the PoLiS (Polygon Line Similarity) metric~\cite{avbelj2014metric} is used to measure the shape difference between the predicted and ground truth polygons, with lower scores indicating higher matching accuracy and better shape alignment. To assess the effectiveness of building polygon extraction, we selected the FFL~\cite{girard2021polygonal}, Hisup~\cite{xu2023hisup}, SAMPolyBuild~\cite{wang2024sampolybuild}, and the recently proposed GCP~\cite{zhang2025global} models, which were trained and tested on the IRSAMap dataset.
\begin{table}[htbp]
\centering
\caption{Results of Road Network Extraction Task on the IRSAMap Dataset}
\resizebox{\columnwidth}{!}{
\begin{tabular}{cccccc}
\hline
\textbf{Method} & \textbf{AP\_poly} & \textbf{AR\_poly} & \textbf{AP\_mask} & \textbf{AR\_mask} & \textbf{Polis} \\ \hline
FFL\textcolor{white}{~\cite{girard2021polygonal}}           & 12.3 & 20.2 & 8.6  & 19.8 & 3.767 \\ 
HiSup\textcolor{white}{~\cite{xu2023hisup} }         & 12.0 & 17.3 & 12.5 & 17.8 & 3.024 \\ 
SAMPolyBuild\textcolor{white}{~\cite{wang2024sampolybuild}}   & 29.9 & 38.5 & 32.7 & 41.9 & 2.560 \\
GCP\textcolor{white}{~\cite{zhang2025global}}   & 30.4 & 43.6 & - & - & 2.218 \\ \hline
\end{tabular}
}
\label{tab:build_results}
\end{table}
Results Analysis:Quantitative precision evaluation results for the four models on the IRSAMap dataset are presented in~\autoref{tab:build_results}. Since the GCP model only outputs polygons, it does not include precision metrics related to masks. As shown in the results, the recently proposed GCP model outperforms the other three models across the three main evaluation metrics. However, despite these improvements, GCP's overall precision still falls short of the requirements for high-precision building vectorization, with both AP and AR failing to exceed 45\%. In the qualitative results shown in ~\autoref{fig:fig7}, the first and sixth columns display the ground truth annotations, which clearly demonstrate the strict labeling strategy of IRSAMap in distinguishing individual buildings—even small, closely adjacent structures are annotated separately. The visualization of FFL in the first column reveals confusion in dense building areas, although its detection performance is slightly better than that of HiSup. Further analysis of~\autoref{fig:fig7} highlights the respective strengths and weaknesses of HiSup and SAMPolyBuild. Specifically, HiSup performs well in building shape regularization and in controlling the number of detected vertices, resulting in simpler and visually cleaner polygon outputs compared to SAMPolyBuild. However, HiSup struggles with distinguishing connected buildings, often misclassifying multiple adjacent buildings as a single instance. In contrast, SAMPolyBuild, which incorporates a post-processing connection algorithm, demonstrates stronger instance separation but also shows some structural confusion in ~\autoref{fig:fig7}. Despite detecting more vertices than HiSup, SAMPolyBuild achieves better performance in identifying individual building entities. This is the main reason for its superior AP and AR scores, as these metrics place greater emphasis on the accurate detection of each separate building instance.Although GCP shows the best qualitative performance overall, its polygon outputs are generated from simplified mask predictions, which introduces certain limitations. For example, in the second row and fifth column of ~\autoref{fig:fig7}, the polygon boundaries appear less smooth, and additional corners can be observed in non-critical areas. These issues stem from the direct influence of mask accuracy on polygon generation. Nevertheless, as seen in the visual results, GCP still demonstrates superior instance detection capability compared to the other models.

\subsection{Large-Scale Land Cover Map}

\begin{figure}[htbp]
  \centering
  \includegraphics[width=\columnwidth]{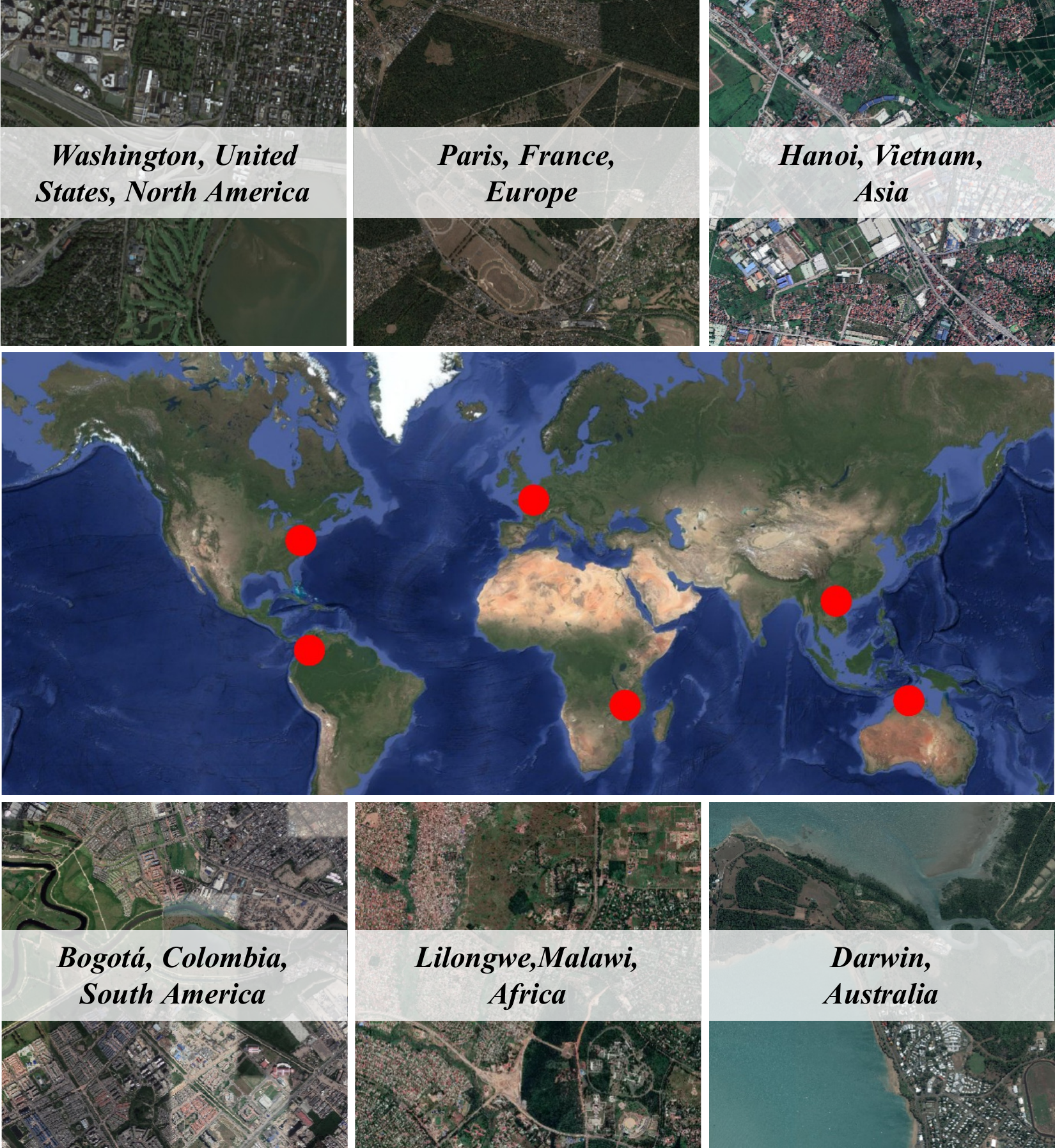} 
  \captionsetup{justification=centering}  
  \caption{Illustration of the selected area for large-scale testing}
  \label{fig:domain_test}
\end{figure}

To assess the generalization capability of IRSAMap for large-scale mapping, representative regions from each of the six continents were selected for evaluation. These test areas were intentionally excluded from all training datasets to ensure an unbiased assessment. As shown in~\autoref{fig:domain_test}, the selected regions exhibit substantial geographic diversity. Inference was performed on these areas using an identical model trained on various datasets, and the prediction results were subsequently analyzed through visual interpretation.

\begin{figure*}[htbp]
  \centering
  \includegraphics[width=\textwidth]{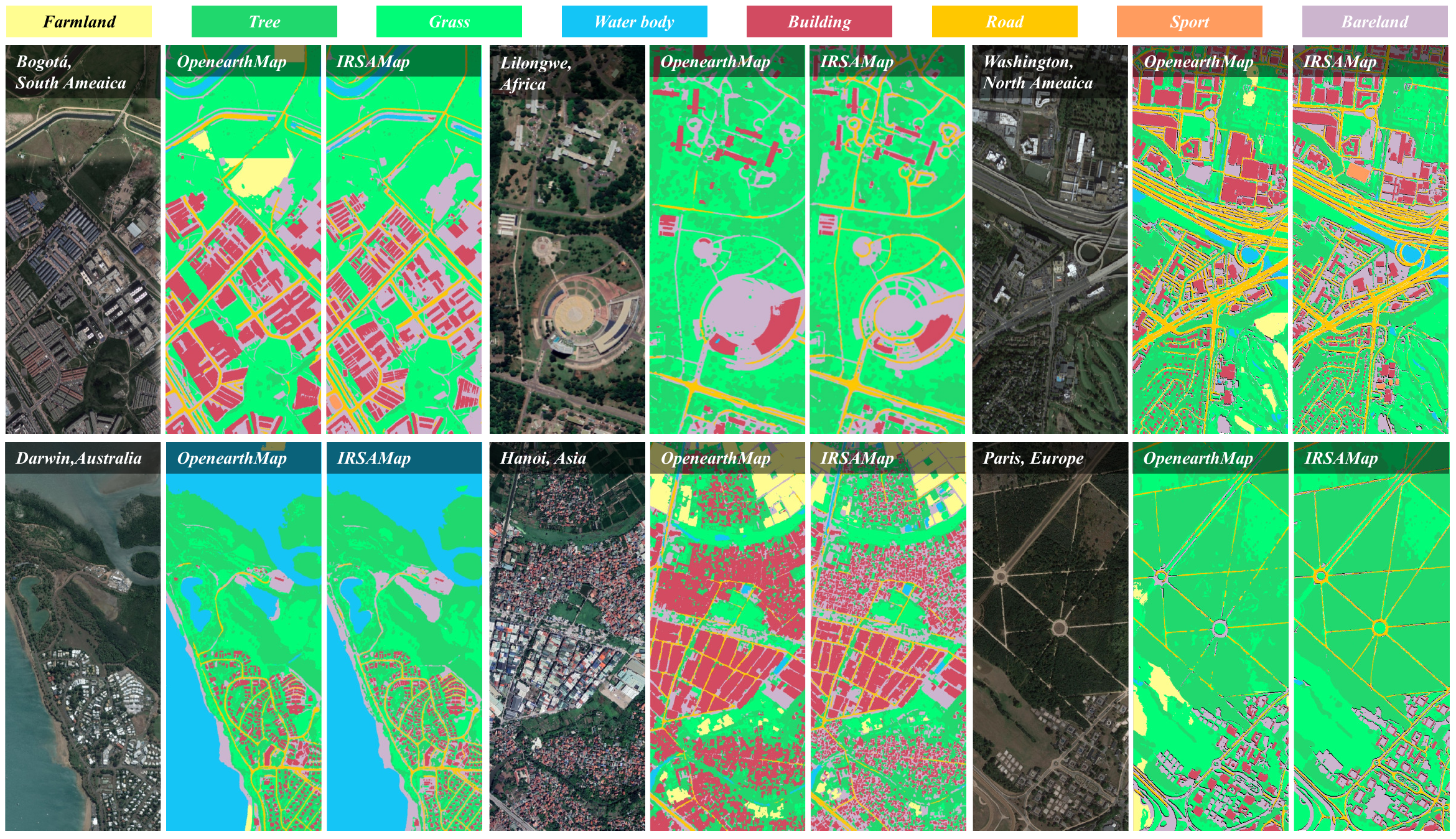} 
  \captionsetup{justification=centering}  
  \caption{Comparison of UPerNet model performance trained on different datasets for land cover classification.}
  \label{fig:domain_results}
\end{figure*}

\begin{figure*}[htbp]
  \centering
  \includegraphics[width=\textwidth]{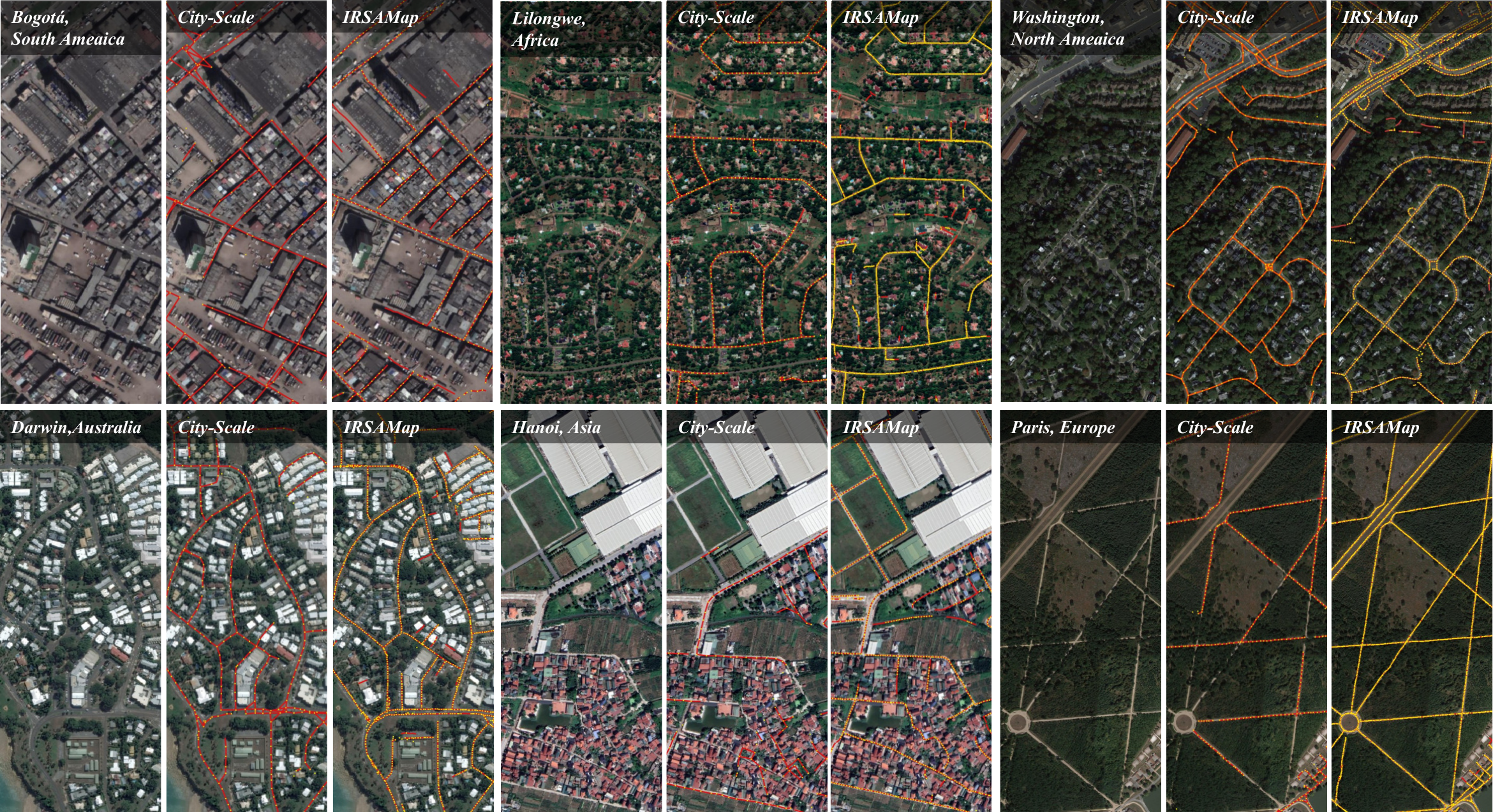} 
  \captionsetup{justification=centering}  
  \caption{Comparison of road network extraction results using the GLD-Road model trained on different datasets.}
  \label{fig:road_domain_test}
\end{figure*}

\autoref{fig:domain_results} shows the visualized land cover classification results of the UPerNet (Swin-S) model trained on different datasets. In multiple regions, the model trained on OpenEarthMap exhibited misclassification in the cropland class. In the Lilongwe region, the model trained on IRSAMap provided more refined vegetation extraction. In addition, since the model based on IRSAMap was trained with roads assigned the highest annotation priority, it demonstrated better road connectivity. For the building class, the model trained on IRSAMap adopted an instance-level annotation strategy, resulting in relatively fragmented outputs but with more precise boundary delineation; in contrast, the model trained on OpenEarthMap tended to represent more complete building units such as individual houses or blocks. Overall, both models exhibited their respective advantages: although OpenEarthMap and IRSAMap had comparable annotation areas (approximately 1000 km²), the consistency in annotation and uniform resolution of IRSAMap enabled its trained model to achieve lower misclassification rates and finer boundary representation across multiple land cover classes.

For the road centerline extraction task, the GLD-Road model was adopted. The Sam-Road model was not selected due to its tendency to produce non-smooth road delineations, which leads to visually unappealing results on large-scale images. The City-Scale dataset features imagery at a 1-meter resolution; thus, all predictions were resampled to 1 m for inference. In contrast, IRSAMap was trained and inferred at a 0.5-meter resolution. This discrepancy in spatial resolution may introduce unfair comparisons in the level of detail captured during road extraction. As illustrated in~\autoref{fig:road_domain_test}, the City-Scale dataset, which contains labels collected exclusively in the United States, presents relatively homogeneous scenes. This results in frequent omissions and discontinuities in the extracted road centerlines. While the current model performs well in scenarios involving single-lane roads, challenges remain in more complex scenes, such as interchanges and multi-lane road networks, where the extracted topologies tend to be disorganized. Notably, the IRSAMap dataset includes annotations for multi-lane roads, indicating that there is still room for improvement in future vectorization models to accurately represent complex road structures.

\begin{figure*}[htbp]
  \centering
  \includegraphics[width=\textwidth]{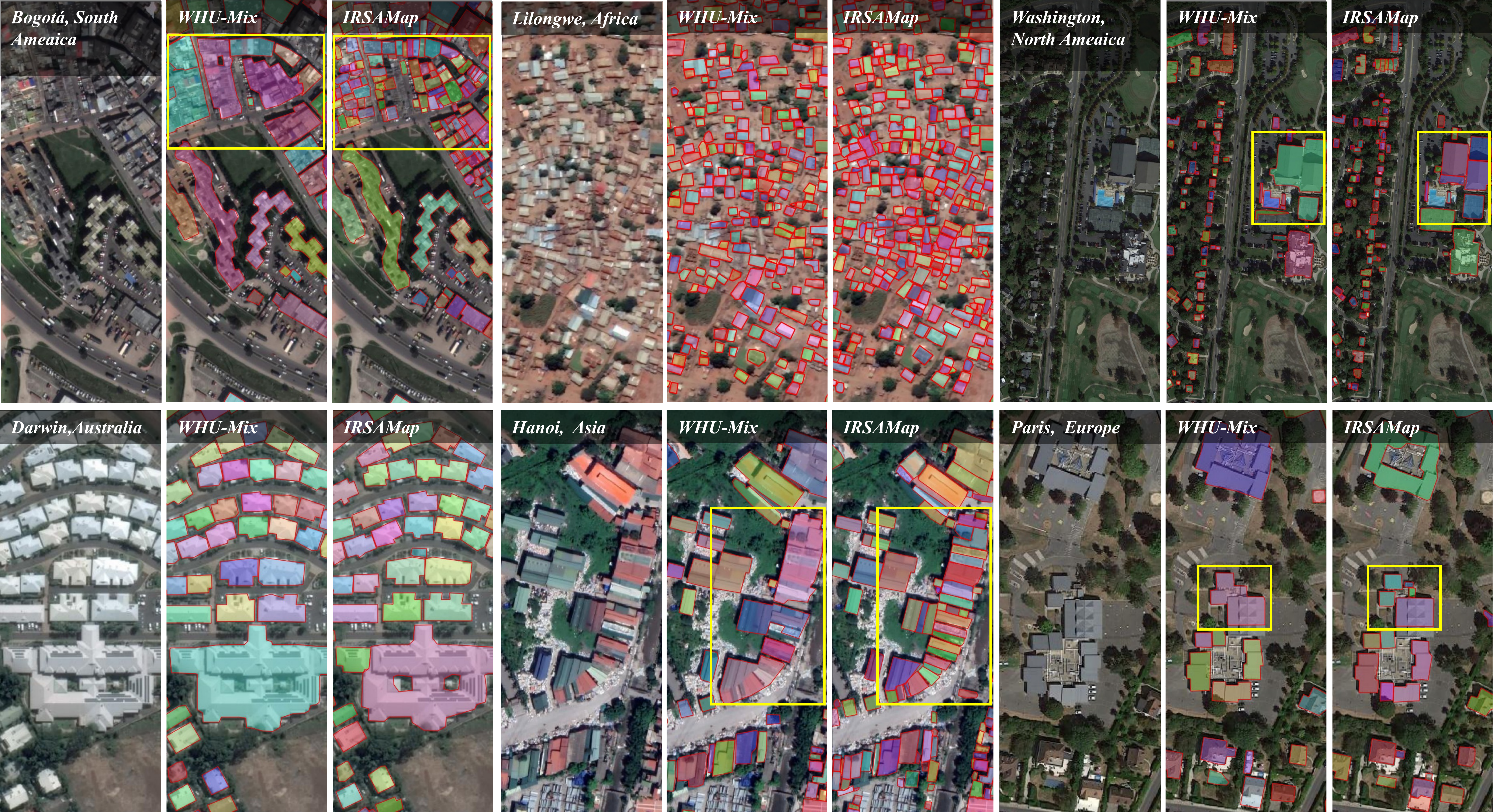} 
  \captionsetup{justification=centering}  
  \caption{Comparison of building extraction results using the GCP model trained on different datasets.}
  \label{fig:building_domain_test}
\end{figure*}


\added[id=R1, comment={7}]{In the task of building footprint extraction, we adopt the GCP model. Compared to the IRSAMap dataset, the WHU-Mix dataset exhibits limitations in the delineation of individual building instances, leading to notable omission errors in densely built-up areas within the generalization domain, as highlighted by the yellow boxes in~\autoref{fig:building_domain_test}. In contrast, both datasets achieve comparable performance in low-density residential areas, such as villa zones and detached houses. Furthermore, most existing high-precision building vectorization methods~\cite{zhang2025global, wang2024sampolybuild, girard2021polygonal} are designed for small, single-frame images, often neglecting the continuity of building outlines across adjacent image tiles. In this section, we address this issue by first applying the GCP model on large-scale images to obtain complete segmentation results, and subsequently introducing a contour simplification module to extract polygonal outlines. This strategy effectively alleviates discontinuities in building boundaries at tile edges.}

\section{Conclusion}
This paper presents IRSAMap, the first large-scale, high-resolution vector dataset for land cover. The dataset includes 10 representative natural and artificial land cover types, all labeled in vector format. IRSAMap offers four key advantages: a comprehensive element vector labeling system, an intelligent labeling process, a multi-task adaptable design, and large-scale data annotation covering 79 typical global regions. It fills a significant gap in the current field of land cover datasets by providing a multi-class vector dataset, making it highly suitable for large-scale vector mapping tasks. Experimental results demonstrate that there is still considerable room for improvement in tasks such as land cover classification, building extraction, and road centerline extraction. This presents opportunities to advance research and applications of vectorization models. IRSAMap provides a standardized benchmark for the pixel-to-object paradigm shift, which will significantly contribute to the development of cutting-edge research areas such as high-precision geospatial feature automation and full-feature collaborative modeling. \added[id=R2, comment={4}]{Although the current IRSAMap is primarily based on high-resolution remote sensing imagery, future versions are planned to incorporate additional multi-modal data sources, such as SAR and hyperspectral data. This will help fill the data gaps in multi-modal remote sensing imagery object vector extraction models and promote research in intelligent remote sensing interpretation and vector extraction models.}


\bibliographystyle{IEEEtran}
\bibliography{irsamapinfer}

\begin{IEEEbiography}[{\includegraphics[width=1in,height=1.25in,clip,keepaspectratio]{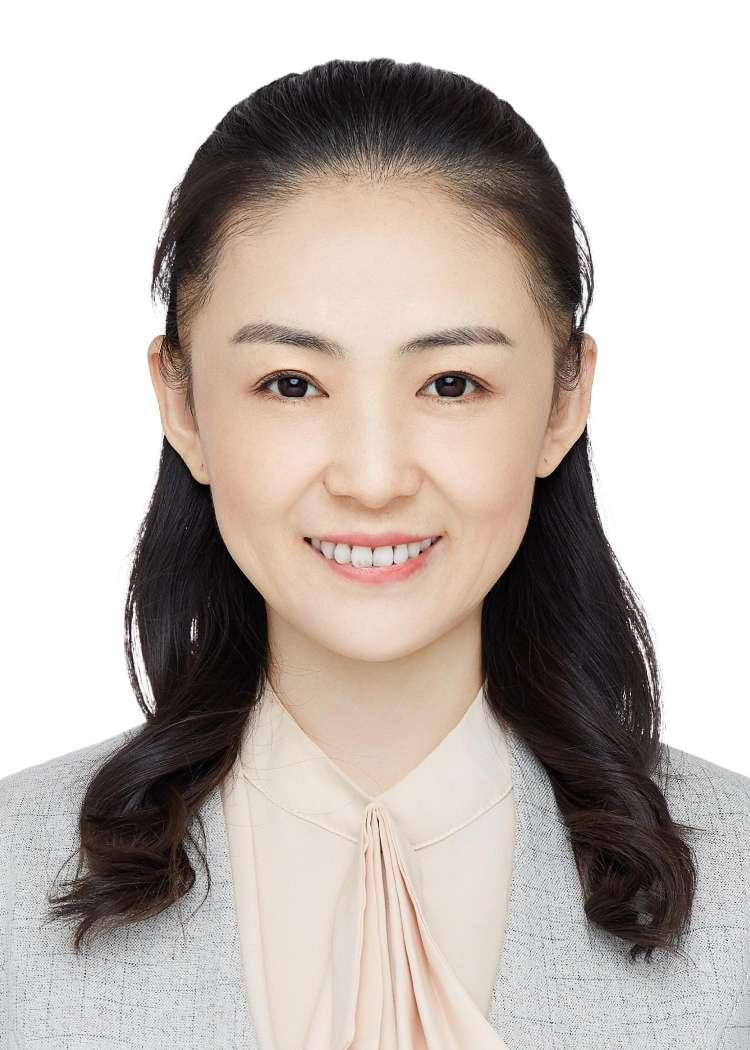}}]{Yu Meng}\\
  received the Ph.D. degree in signal and information processing from the Institute of Remote Sensing Applications, Chinese Academy of Sci-ences, Beijing, China, in 2008.\\
  \indent She is currently a professor at the Aerospace Information Research Institute, Chinese Academy of Sciences. Her research interests include intelligent interpretation of remote sensing images,remote sensing time-series signal processing, and big spatial-temporal data application.\\
  \indent Dr Meng serves as an editor and board member of the National Remote Sensing Bulletin, Journal of Image and Graphics.

\end{IEEEbiography}

\begin{IEEEbiography}[{\includegraphics[width=1in,height=1.25in,clip,keepaspectratio]{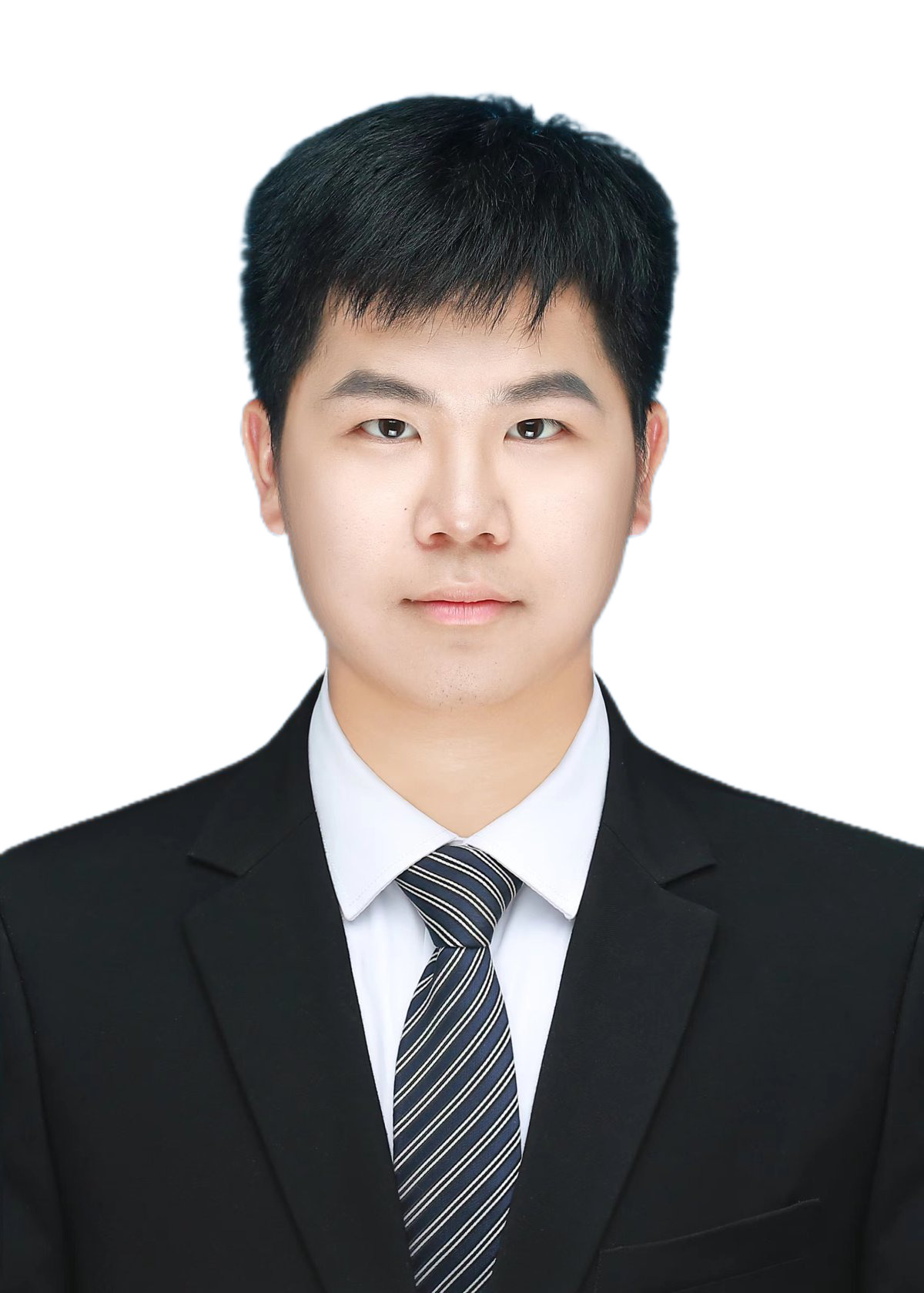}}]{Ligao Deng}\\
  received the B.S. degree in automation from the College of Mechanical and Electrical Engineering, Northeast Forestry University, Harbin, China, in 2022. He is currently pursuing the Ph.D. degree in signal and information processing with the Aerospace Information Research Institute, University of Chinese Academy of Sciences, Beijing, China. His research interests include computer vision and intelligent interpretation of remote sensing imagery.
\end{IEEEbiography}

\begin{IEEEbiography}[{\includegraphics[width=1in,height=1.25in,clip,keepaspectratio]{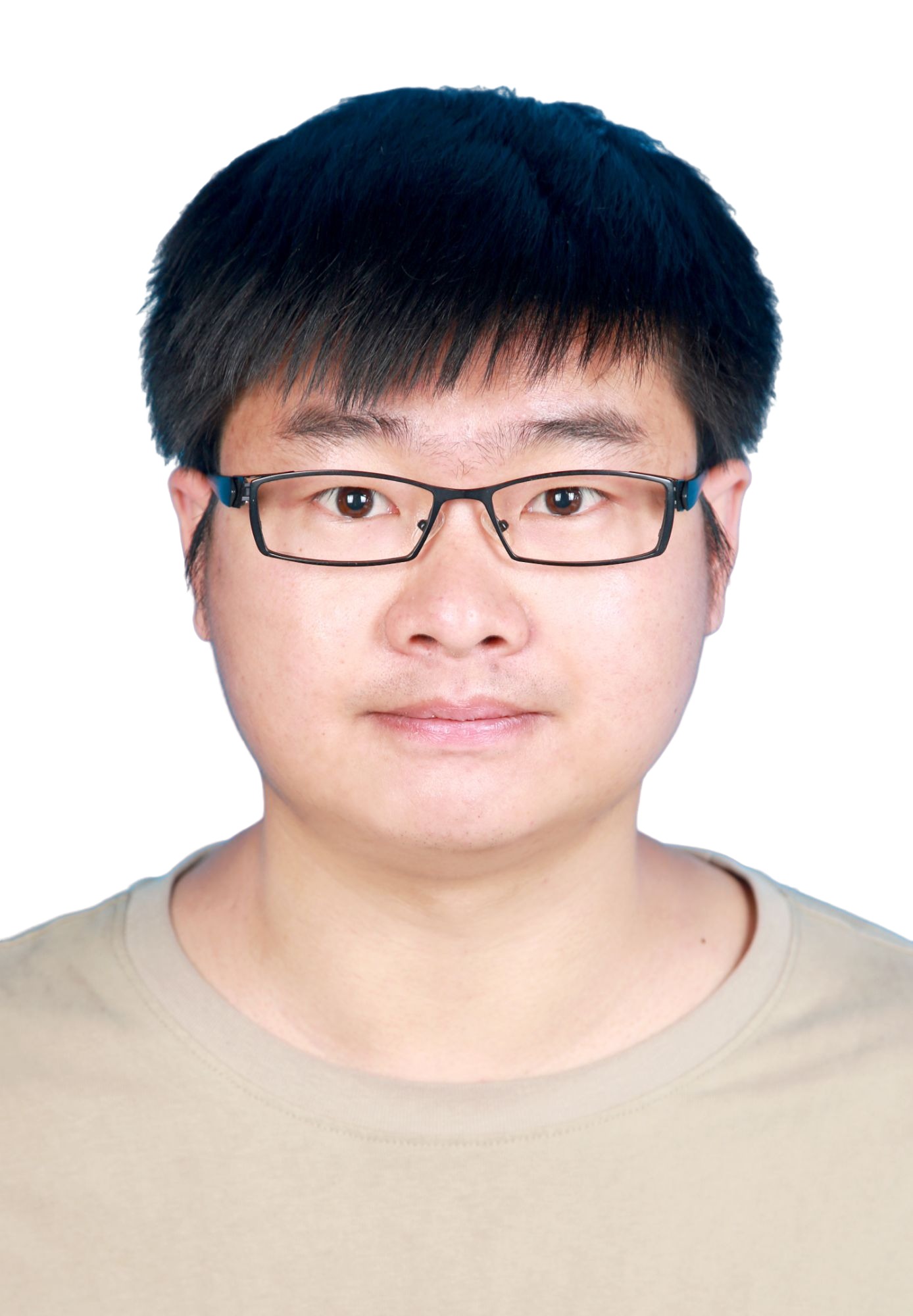}}]{Zhihao Xi}\\
  received the B.S. degree from the Wuhan University of Technology, Wuhan, China, in 2019, and the Ph.D. degree from the Aerospace Information Research Institute, Chinese Academy of Sciences (CAS), Beijing, China, in 2024.
 He is currently an Assistant Professor with the Aerospace Information Research Institute, CAS. His research interests include computer vision, domain adaptation, and remote sensing image interpretation.
\end{IEEEbiography}

\begin{IEEEbiography}[{\includegraphics[width=1in,height=1.25in,clip,keepaspectratio]{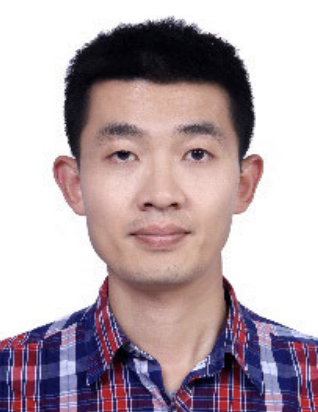}}]{Jiansheng Chen}\\received the Ph.D. degree from
  the Institute of Remote Sensing and Digital Earth,
  Chinese Academy of Sciences, Beijing, China,
  in 2012.\\
  \indent He is currently an Assistant Professor with the
  Aerospace Information Research Institute, Chinese
  Academy of Sciences. His research interests
  include object detection and remote sensing image
  processing.
  \end{IEEEbiography}

\begin{IEEEbiography}[{\includegraphics[width=1in,height=1.25in,clip,keepaspectratio]{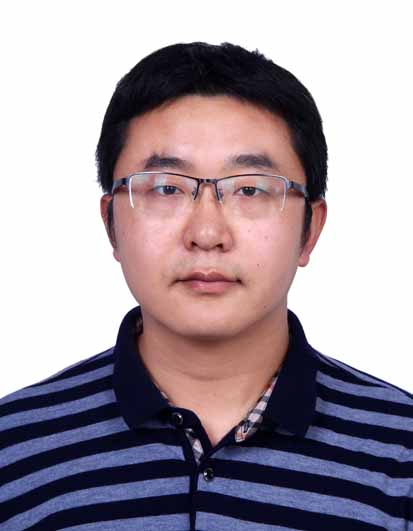}}]{Jingbo Chen}\\
  received the Ph.D.degree in cartography and geographic information systems from the Institute of Remote Sensing Applications, Chinese Academy of Sciences,Beijing, China, in 2011.\\
  \indent He is currently an Associate Professor with the Aerospace Information Research Institute, Chinese Academy of Sciences. His research interests cover intelligent remote sensing analysis, integrated application of communication, navigation, and remote sensing.
  \end{IEEEbiography}

\begin{IEEEbiography}[{\includegraphics[width=1in,height=1.25in,clip,keepaspectratio]{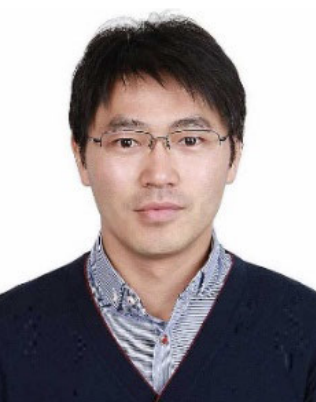}}]{Anzhi Yue}\\
  received the Ph.D. degree in agricultural information technology from China Agricultural University, Beijing, China, in 2012.\\
  \indent He is currently a Professor with the Aerospace
  Information Research Institute, Chinese Academy of
  Sciences, Beijing. His research interests include the
  fields of land use and land cover classification, with
  a focus on remote sensing technologies and their
  application in land survey and monitoring.
  \end{IEEEbiography}

\begin{IEEEbiography}[{\includegraphics[width=1in,height=1.25in,clip,keepaspectratio]{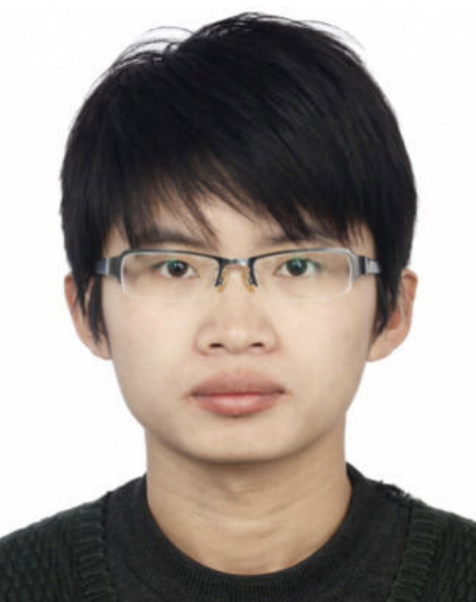}}]{Diyou Liu}\\
  received the Ph.D. degree in agricultural
  information technology from the College of Land
  Science and Technology, China Agricultural University, Beijing, China, in 2021.\\
  \indent He is currently a Special Research Assistant at the
  Aerospace Information Research Institute, Chinese
  Academy of Sciences, Beijing. His research focuses
  on the production of cartographic-level vector element data using intelligent interpretation methods
  from high-resolution remote sensing imagery.
  \end{IEEEbiography}

\begin{IEEEbiography}[{\includegraphics[width=1in,height=1.25in,clip,keepaspectratio]{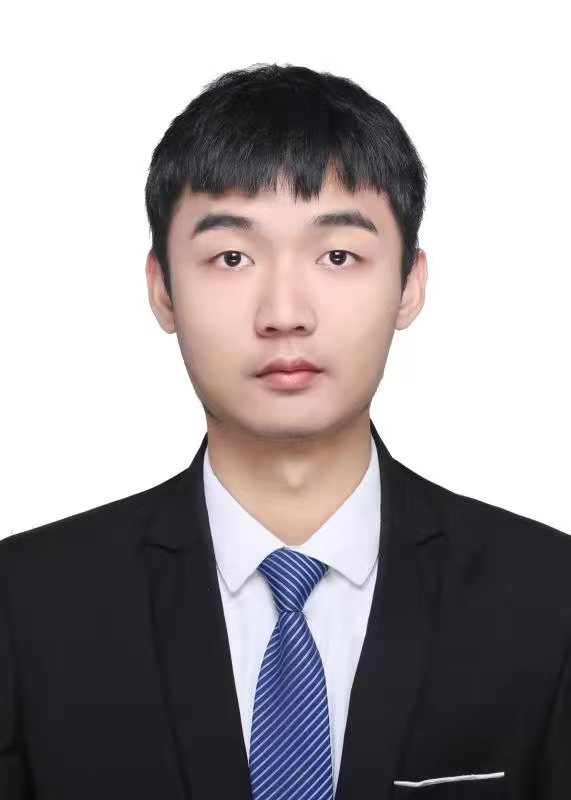}}]{Kai Li}\\
      received a bachelor‘s degree in engineering, spatial information and digital technology from UESTC, Chengdu, China, in 2021. He is currently pursuing a PhD degree with UCAS, Beijing, China, supervised by Zhongming Zhao and Yu Meng. His research interests include remote sensing, computer vision, and machine learning.
  \end{IEEEbiography}

\begin{IEEEbiography}[{\includegraphics[width=1in,height=1.25in,clip,keepaspectratio]{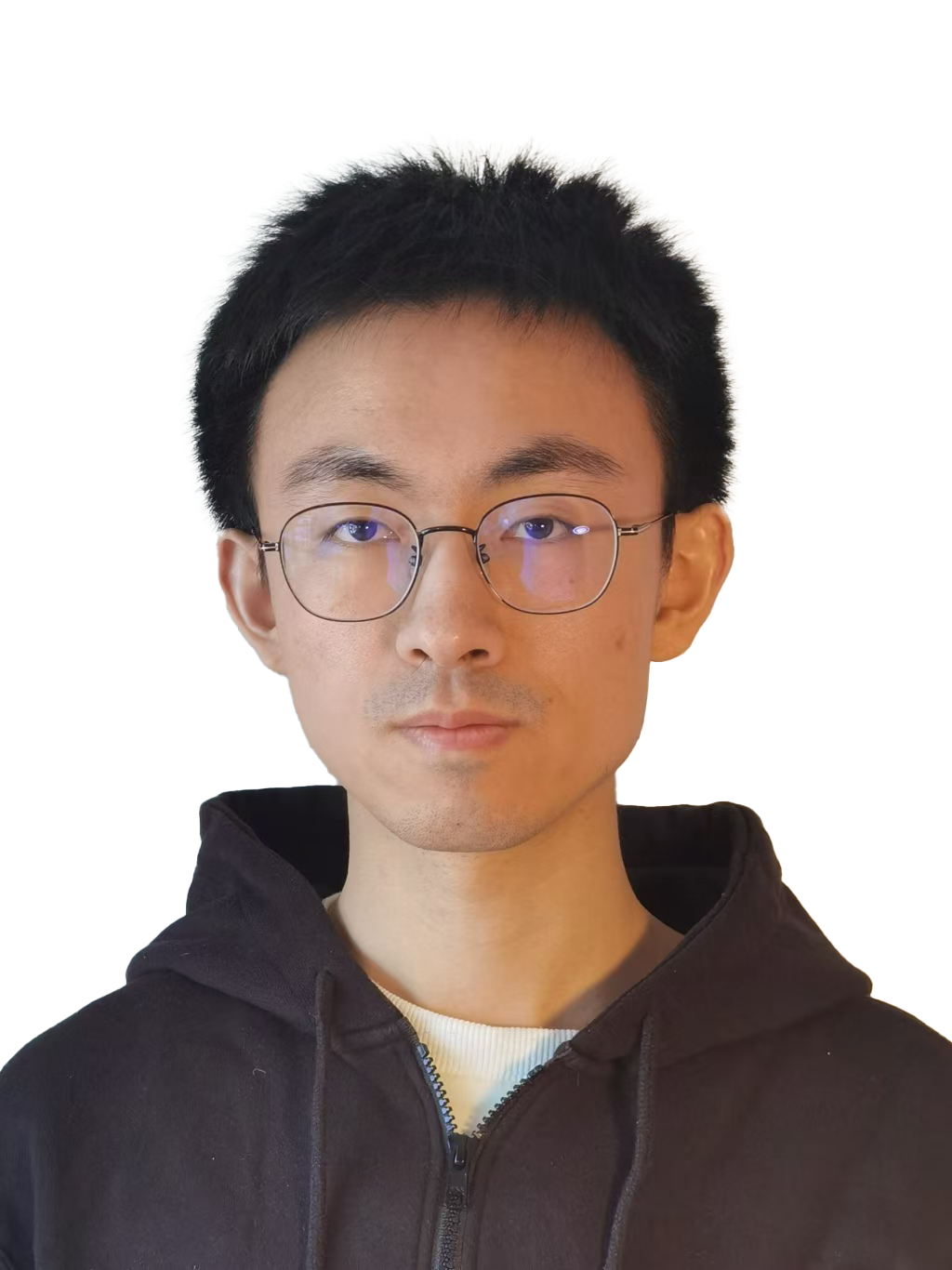}}]{Chenhao Wang}\\
    received the bachelor’s degree from the University of Electronic Science and Technology of China, Chengdu, China, in 2022. He is currently pursuing the Ph.D. degree with the University of Chinese Academy of Sciences, Beijing, China.His research focuses on building extraction from remote sensing images.
 \end{IEEEbiography}

\begin{IEEEbiography}   
  [{\includegraphics[width=1in,height=1.25in,clip,keepaspectratio]{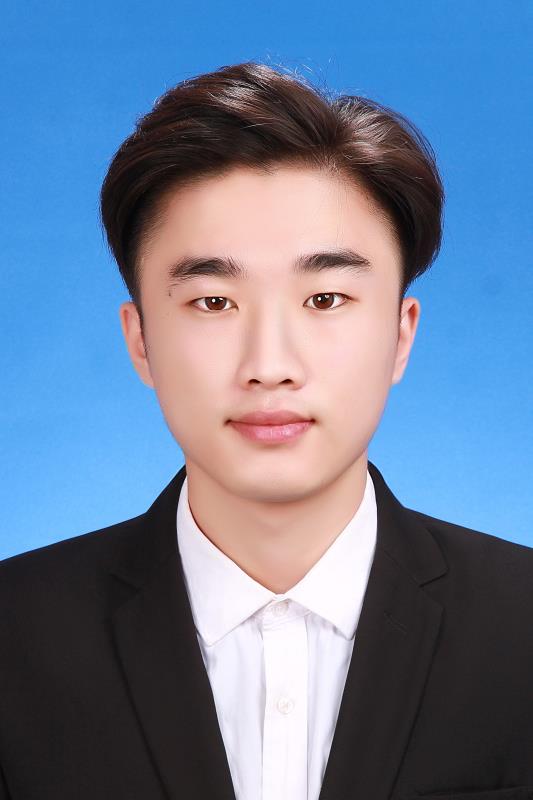}}]{Kaiyu Li}\\    received the B.E. and M.E. degrees from Shandong University of Science and Technology, Qingdao, China, in 2020 and 2023, respectively. He is currently pursuing the Ph.D. degree with the School of Software Engineering, Xi'an Jiaotong University, Xi'an, China. His research interests include image processing and remote-sensing visual perception.
\end{IEEEbiography}

\begin{IEEEbiography}
  [{\includegraphics[width=1in,height=1.25in,clip,keepaspectratio]{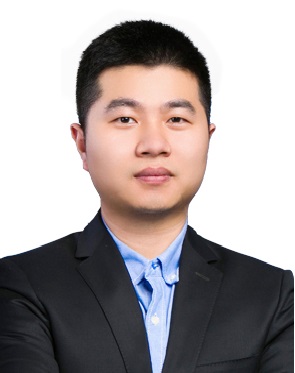}}]{Yupeng Deng}\\
  received the Ph.D. degree from the Aerospace Information Research Institute, Chinese Academy of Sciences, Beijing, China, in 2023.Now, he is a Post-Doctoral Researcher Supervised by Jianhua Gong. He is specialized in remote sensing and change detection.
\end{IEEEbiography}

\begin{IEEEbiography}
  [{\includegraphics[width=1in,height=1.25in,clip,keepaspectratio]{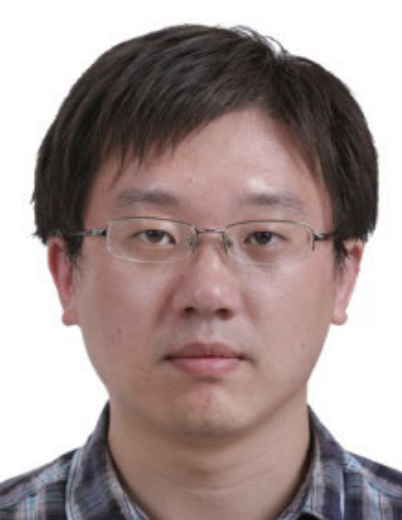}}]{Xian Sun}\\
  (Senior Member, IEEE) received the B.Sc.
  degree from Beijing University of Aeronautics and
  Astronautics, Beijing, China, in 2004, and the M.Sc.
  and Ph.D. degrees from the Institute of Electronics,
  Chinese Academy of Sciences, Beijing, in 2009.
  \indent He is currently a Professor with the Aerospace
  Information Research Institute, Chinese Academy
  of Sciences. His research interests include computer
  vision, geospatial data mining, and remote sensing
  image understanding.
\end{IEEEbiography}

\end{document}